\documentclass[12pt]{elsarticle}
\usepackage[utf8]{inputenc}
\usepackage{setspace}
\usepackage[margin=0.75in]{geometry}
\usepackage{graphicx}
\graphicspath{ {./figures/} }
\usepackage[labelfont=bf, font = normalsize]{caption}
\usepackage{subcaption}
\usepackage{amsmath,amssymb}
\usepackage{mathrsfs}
\usepackage{bm}
\usepackage{xcolor}
\usepackage{array}
\usepackage{appendix}
\usepackage{tikz}
\usepackage{floatrow}
\usepackage{rotating}
\usepackage{multirow,bigdelim}
\usepackage{array, blkarray, makecell}
\usepackage{tabularx}
\DeclareFloatFont{small}{\footnotesize}
\newcolumntype{P}[1]{>{\centering\arraybackslash}p{#1}}

\providecommand{\keywords}[1]
{
  \small	
  \textbf{\textit{Keywords:}} #1
}

\usepackage{siunitx}
\usepackage{tikz}
\usepackage{tikz-3dplot}
\usepackage{graphicx}
\usetikzlibrary{calc,backgrounds}
\usepackage{pgfplotstable}
\usepgfplotslibrary{groupplots}
\usetikzlibrary{intersections}
\usepgfplotslibrary{fillbetween}

\definecolor{rwth1}{RGB}{0,84,159}      
\definecolor{rwth2}{RGB}{142,186,229}   
\definecolor{rwth3}{RGB}{0,97,101}      
\definecolor{rwth4}{RGB}{0,152,161}     
\definecolor{rwth5}{RGB}{87,171,39}     
\definecolor{rwth6}{RGB}{189,205,0}     
\definecolor{rwth7}{RGB}{255,237,0}     
\definecolor{rwth8}{RGB}{246,168,0}     
\definecolor{rwth9}{RGB}{227,0,102}     
\definecolor{rwth10}{RGB}{204,7,30}     
\definecolor{rwth11}{RGB}{161,16,53}    
\definecolor{rwth12}{RGB}{97,33,88}     
\definecolor{rwth13}{RGB}{122,111,172}  

\definecolor{rwthb1}{HTML}{e8f1fa}      
\definecolor{rwthb2}{HTML}{c7ddf2}      
\definecolor{rwthb3}{HTML}{8ebae5}      
\definecolor{rwthb4}{HTML}{407fb7}      
\definecolor{rwthb5}{HTML}{00549f}      

\definecolor{rwtho1}{HTML}{fff7ea}      
\definecolor{rwtho2}{HTML}{feeac9}      
\definecolor{rwtho3}{HTML}{fdd48f}      
\definecolor{rwtho4}{HTML}{fabe50}      
\definecolor{rwtho5}{HTML}{f6a800}      

\tikzstyle{dashpattern0} = [dash pattern = ]
\tikzstyle{dashpattern1} = [dash pattern = on 4.25pt off 0.75pt]
\tikzstyle{dashpattern2} = [dash pattern = on 1.5pt off 0.5pt]
\tikzstyle{dashpattern3} = [dash pattern = on 0.75pt off 0.4pt]
\tikzstyle{dashpattern4} = [dash pattern = on 3pt off 1pt on 1pt off 1pt]
\tikzstyle{dashpattern5} = [dash pattern = on 3.75pt off 0.5pt on 0.75pt off 0.5pt on 0.75pt off 0.5pt]
\tikzstyle{dashpattern6} = [dash pattern = on 3.25pt off 0.5pt on 0.75pt off 0.5pt on 0.75pt off 0.5pt on 0.75pt off 0.5pt]
\tikzstyle{dashpattern7} = [dash pattern = on 3.25pt off 0.5pt on 0.75pt off 0.5pt on 0.75pt off 0.5pt on 0.75pt off 0.5pt on 0.75pt off 0.5pt]
\tikzstyle{dashpattern8} = [line cap=round, dash pattern = on 3.25pt off 2.75pt]
\tikzstyle{dashpattern9} = [line cap=round, dash pattern = on 0.01pt off 2pt]
\tikzstyle{dashpattern10}= [line cap=round, dash pattern = on 3.25pt off 2pt on 0.01pt off 2pt]
\tikzstyle{dashpattern11}= [line cap=round, dash pattern = on 3.5pt off 1.75pt on 0.01pt off 1.75pt on 0.01pt off 1.75pt]
\tikzstyle{dashpattern12}= [line cap=round, dash pattern = on 3.5pt off 1.75pt on 0.01pt off 1.75pt on 0.01pt off 1.75pt on 0.01pt off 1.75pt]
\tikzstyle{dashpattern13}= [line cap=round, dash pattern = on 3.5pt off 1.75pt on 0.01pt off 1.75pt on 0.01pt off 1.75pt on 0.01pt off 1.75pt on 0.01pt off 1.75pt]
            
\usepackage{tablefootnote}    
\usepackage{algorithm}
\usepackage{algpseudocode}

\usepackage{hyperref}

\usepackage{amsthm}

\newtheorem{remark}{Remark}  
            
\usepackage{natbib}
\title{%
	\LARGE
	Automated Model Discovery for Tensional Homeostasis: \large Constitutive Machine Learning in Growth and Remodeling}

\author[1]{Hagen Holthusen \corref{cor1}}
\ead{hagen.holthusen@ifam.rwth-aachen.de}
\cortext[cor1]{Corresponding author\\hagen.holthusen@ifam.rwth-aachen.de\\ Mies-van-der-Rohe-Str. 1, 52074 Aachen, Germany}
\author[1]{Tim Brepols}
\author[1,2]{Kevin Linka}
\author[3]{Ellen Kuhl}

\address[1]{Institute of Applied Mechanics, RWTH Aachen University, Germany}
\address[2]{Institute for Continuum and Material Mechanics, Hamburg University of Technology, Germany}
\address[3]{Department of Mechanical Engineering, Stanford University, United States}

\date{}

\begin{document}


\begin{abstract}
Soft biological tissues exhibit a tendency to maintain a preferred state of tensile stress, known as tensional homeostasis, which is restored even after external mechanical stimuli. This macroscopic behavior can be described using the theory of kinematic growth, where the deformation gradient is multiplicatively decomposed into an elastic part and a part related to growth and remodeling. Recently, the concept of homeostatic surfaces was introduced to define the state of homeostasis and the evolution equations for inelastic deformations.

However, identifying the optimal model and material parameters to accurately capture the macroscopic behavior of inelastic materials can only be accomplished with significant expertise, is often time-consuming, and prone to error, regardless of the specific inelastic phenomenon. To address this challenge, built-in physics machine learning algorithms offer significant potential.

In this work, we extend our inelastic Constitutive Artificial Neural Networks (iCANNs) by incorporating kinematic growth and homeostatic surfaces to discover the scalar model equations, namely the Helmholtz free energy and the pseudo potential. The latter describes the state of homeostasis in a smeared sense.
We evaluate the ability of the proposed network to learn from experimentally obtained tissue equivalent data at the material point level, assess its predictive accuracy beyond the training regime, and discuss its current limitations when applied at the structural level.

Our source code, data, examples, and an implementation of the corresponding material subroutine are made accessible to the public at \url{https://doi.org/10.5281/zenodo.13946282}.\\
\\
\end{abstract}

\maketitle


\section{Introduction}
\label{sec:intro}
Patient healthcare increasingly focuses on personalizing treatments to meet the unique needs of each individual. However, the complex mechanisms and mechanical behaviors of biological systems make this task particularly challenging. Despite major advancements in modeling biomechanical behavior over recent decades, identifying patient-specific parameters remains a significant hurdle. Data-driven models offer a promising solution by leveraging patient data to pinpoint these individualized characteristics. Moreover, unbiased generic models avoid assuming specific behaviors a priori, enhancing our understanding of tissue behaviors by unveiling underlying mechanisms. This integration of data-driven tools is accelerating the development of more precise, patient-specific therapies across diverse medical fields.

In this context, testing the mechanical behavior of both living and non-living materials is notoriously labor-intensive, leading to limited data availability. This poses a significant challenge for standard neural networks, which struggle to learn from small datasets and often produce unphysical predictions outside their training range. However, the inherent adherence of living tissues to fundamental physical laws presents an exciting opportunity. By embedding physics directly into neural networks, we can drastically reduce data requirements while enabling the networks to autonomously discover complex, biologically accurate models. These biologically accurate models go beyond traditional simulations by incorporating key characteristics, such as tensional homeostasis, which are critical for capturing the true behavior of biological systems.

These diverse physics-embedded approaches promise to revolutionize how we model biological systems, combining the strengths of physics and machine learning to achieve unprecedented accuracy and efficiency.

\subsection{State-of-the-art}

\textbf{Data-driven constitutive modeling.}
The integration of expert knowledge, particularly in continuum mechanics, with the computational power of data-driven models enables the development of highly efficient numerical frameworks. These architectures can discover, and most importantly, predict physically plausible material behaviors beyond the training range. Broadly, we can distinguish between \textit{model-free} approaches \cite{kirchdoerfer2016,eggersmann2021}, and those that enforce physical laws either \textit{strongly} or \textit{weakly}. The model-free approach operates without constitutive assumptions, while the latter typically incorporates physics during training, for instance, by modifying the loss function as in physics-informed neural networks \cite{raissi2018}.

Numerous robust methods for \textit{strongly enforcing} physical laws have emerged recently. In this contribution, we present a neural network architecture that falls within this category, focusing specifically on data-driven constitutive modeling — teaching constitutive knowledge to the data-driven framework of interest.
We kindly refer to the contribution by \citet{linden2023} for a more theoretical and in-depth discussion on this topic.

In recent years, most strategies for integrating thermodynamics into neural networks initially focused on small strains and elasticity, gradually extending to inelastic effects. For example, Thermodynamics-based Artificial Neural Networks were introduced by \cite{masi2021,masi2023}, and Mechanics-Informed Artificial Neural Networks \cite{asad2022} were later expanded to account for visco-elasticity \cite{asad2023}. A key principle in these frameworks is the omnipresence of convexity in constitutive equations, which is leveraged in models like input convex neural networks when incorporating physical laws. One notable example is the physics-augmented Neural Network \cite{klein2023}.

To address visco-elastic and more complex inelastic phenomena, generalized standard materials approaches can be adopted, incorporating a dissipation potential tied to inelasticity rates \cite{rosenkranz2024}. Similarly, unsupervised material discovery frameworks, such as EUCLID \cite{flaschel2021}, have been expanded to automatically identify inelastic behaviors in small strains \cite{flaschel2022,flaschel2023}.
Further, physics-informed constitutive models trained through probabilistic machine learning for isotropic and anisotropic hyperelasticity were proposed \cite{fuhg2022} and applied to discover the response of strain-rate-sensitive materials \cite{upadhyay2024}.

In contrast to physics-augmented neural networks, we follow the Constitutive Artificial Neural Networks (CANNs) approach \cite{linka2021,linka2023}, where the entire architecture is explicitly custom-designed to adhere to thermodynamic principles. Like EUCLID, CANNs produce generalizable and mechanically interpretable neural networks, which offer insights into the material properties embedded in experimental data. 
In a biomedical context, CANNs were used to identify and study the mechanical behavior of skin \cite{linka2023b} and the human brain \cite{LinkaPierreEtAl2023}.
CANNs have been further extended to model visco-elastic behavior through a Prony-series approach \cite{AbdolaziziLinkaEtAl2023,wang2023}, and general inelasticity at finite strains and strain rates with the inelastic Constitutive Artificial Neural Network (iCANN) \cite{holthusen2024,boes2024arxiv}.

This literature review represents only a small fraction of the current trends in the mechanics community's focus on data-driven models. It is by no means exhaustive, particularly given the high level of activity in this area, with numerous leading experts and prominent research groups contributing to its rapid development. For a more comprehensive overview of the diverse approaches being explored, we refer readers to the survey by \citet{watson2024arxiv}.

Standard artificial neural networks tend to be densely connected, making it difficult to interpret the relationships between inputs and outputs. However, in constitutive modeling, \textit{sparsity} is crucial. Simpler models that effectively explain the data are preferred over non-interpretable ones and, within reasonable limits, can be considered unique. This focus on sparsity enhances the model's interpretability and aligns with the goal of deriving meaningful insights from the underlying physical processes.
This becomes even more critical when the data is subject to uncertainties \cite{linka2024arxiv}, where achieving a deterministic relationship between stresses and strains is often unrealistic. During network training, various regularization techniques can be applied to help identify a meaningful subset of potential models from the array of approaches available \cite{mcculloch2024}. A related strategy involves starting with a simplified subset of the original neural network, ideally beginning with a `one-term model'. The complexity of the network is then progressively increased by iteratively incorporating additional terms \cite{linka2024}. This approach facilitates control over the model’s complexity, promoting both interpretability and robust generalization.

Once a model has been discovered to capture the relationship between stresses and strains at the material point level, the next step is to explore its predictive power on a structural scale. Since all of the aforementioned approaches are rooted in thermodynamic principles, they can be seamlessly integrated into numerical boundary value solvers like Finite Element Analysis.
In the biomedical field, this has been achieved with notable success, enabling the simulation of hyperelastic behavior in complex biological structures, such as the human brain \cite{peirlinck2024}, the aortic arch \cite{peirlinck2024b}, and cardiac tissues \cite{martanova2024}. These advancements not only push the boundaries of material modeling but also pave the way for more accurate, patient-specific simulations in medical applications.\newline

\textbf{Tensional homeostasis.} Biological tissues are characterized by their ability to actively respond to external mechanical stimuli \cite{wolff1870}, a response driven by complex interactions at the cellular level. These interactions give rise to macroscopic observations such as growth and remodeling, which, rather than being directly traced to micromechanical mechanisms, are treated in a phenomenological, large-scale sense. For detailed microscale descriptions, see \cite{silver2003, cowin2007}. From a mechanical standpoint, roughly five modeling approaches in continuum mechanics address these active deformations. Readers are referred to the insightful article by \citet{goriely20175ways} for further details. In the context of growth and remodeling, the proposed iCANN framework follows the active strain approach.

Hard and soft tissues can be broadly distinguished, both being extensively studied through numerical and experimental approaches. Hard tissues, such as bones \cite{cowin2001, floerkemeier2010} and dental tissues \cite{wilmers2024}, and soft tissues, including brain \cite{budday2020, reiter2021, reiter2023PAMM}, arteries \cite{holzapfel2000, holzapfel2010} including considerations of the multifield problem \cite{manjunatha2022}, the cardiovascular system \cite{humphrey2002book,hoskins2017,quarteroni2017}, and the human heart \cite{guelcher2023, dal2013} with its complex multiphysical interactions \cite{tikenogullari2024} and its maturation process in biohybrid heart valve implants \cite{sesa2023}, remain active areas of research. This wide scope reflects the need to better understand both underlying mechanisms and macroscopic behavior. While hard tissues primarily experience finite rotations, soft tissues are subject to both finite strains and rotations.

Living tissues actively adapt to their environment, with one key feature being homeostasis—the ability to achieve, maintain, and restore a stable mechanical state under changing physiological conditions \cite{humphrey2014, cyron2016, cyron2017}. \citet{brown1999} introduced the hypothesis of \textit{tensional homeostasis}, suggesting that soft tissues prefer a tensile stress state. However, achieving this state without external loading is constrained by the availability of hormones and nutrients. One of the few continuum models incorporating these multiphysical interactions is presented in \cite{soleimani2020}.

Recent experimental studies have provided insights into tensional homeostasis. For example, \cite{marenzana2006, ezra2010} performed uniaxial tests on tissue equivalents, observing active contractions and restoration of the homeostatic state. Since real tissues typically experience multiaxial loading, \cite{eichinger2020, eichinger2021b, eichinger2021c} extended these studies to multiaxial scenarios, including non-proportional load perturbations of the state of homeostasis. They observed that tissue equivalents maintain and restore the state of homeostasis even for such more sophisticated loadings. More recently, \citet{paukner2024} used this setup to investigate vascular smooth muscle cell-seeded tissue equivalents under dynamic biaxial loading. For a comprehensive review, see \citet{eichinger2021}.

The two most common approaches for modeling growth and remodeling are constrained mixture models, introduced by \citet{humphrey2002}, and kinematic growth models \cite{rodriguez1994}. This contribution follows the kinematic approach, which multiplicatively decomposes the deformation gradient. According to \cite{cyron2016, braeu2017}, the inelastic part might be further decomposed into growth and remodeling components to better distinguish these processes of (volumetric) growth and remodeling.

Growth and remodeling models can be classified as \textit{isotropic} \cite{lubarda2002, himpel2005, kuhl2007} or \textit{anisotropic} \cite{goriely2017, zahn2018, rahman2023}. Anisotropic models describe induced anisotropic behavior resulting from growth and remodeling processes, meaning that a material, which might be initially isotropic, develops directional behavior due to these processes. \citet{braeu2017} observed a general mismatch between isotropic growth models and experimental observations, emphasizing the need for anisotropic models to capture realistic tissue behavior.

\subsection{Hypothesis}
Various physics-based networks have been proposed to model material behavior, with some extensions for inelastic materials, though few apply to the finite strain regime. While early applications of such networks to tissues, such as modeling the visco-elastic behavior of brains \cite{hinrichsen2024}, exist, networks for growth and remodeling remain largely unexplored. 
To the authors' knowledge, no approach has yet been proposed to design physics-embedded networks for discovering tensional homeostasis.

This contribution hypothesizes that by enhancing the inelastic Constitutive Artificial Neural Network (iCANN) framework \cite{holthusen2024,holthusen2024PAMM} for growth and remodeling, using the concept of homeostatic surfaces \cite{lamm2022,holthusen2023,holthusen2023PAMM}, it is possible to uncover the mechanisms of tensional homeostasis.

At this stage of development, we neglect any directional dependence and assume the tissue to behave initially isotropic.
We improve the original iCANN by focusing on network architecture design, ensuring satisfaction of equality constraints, and formulating a general approach to capture rate-dependent behavior. 
The feed-forward network of the pseudo potential is expressed in terms of principal stresses, with refined activation functions that advance the networks’s ability to capture complex material responses.

\subsection{Outline}
In Section~\ref{sec:const}, we review the thermodynamically consistent constitutive framework for inelastic materials at finite strains, highlighting the physical restrictions of the Helmholtz free energy and the concept of homeostatic surfaces/pseudo potentials in Section~\ref{sec:potential}. 
We then incorporate these considerations into the iCANN architecture in Section~\ref{sec:network}, employing a time-discretized approach, iteratively solve the nonlinear homeostasis equation and present our enhanced designs for the energy and potential.
In Section~\ref{sec:results}, we train our network using uniaxial and biaxial experimental data of tissue equivalents and evaluate its performance.
We critically discuss the limitations encountered during these observations in Section~\ref{sec:limits}. 
Finally, we conclude our findings and provide an outlook in Section~\ref{sec:outlook}.

\section{Constitutive framework}
\label{sec:const}
In the following, we briefly recapitulate the constitutive framework of inelastic Constitutive Artificial Neural Networks presented by \citet{holthusen2024} and its polyconvex extension \cite{holthusen2024PAMM} in order to specialize it to account for growth and remodeling.
A more detailed discussion of the underlying concepts can be found in there.

\subsection{Governing equations}
\textbf{Kinematics.} We describe the deformation as well as the material's response of a body by the deformation gradient $\bm{F}$.
Consequently, a rigid motion of both the reference configuration, $\bm{F}^\# = \bm{F}\bm{Q}^{\#^T}$, and the current configuration, $\bm{F}^+=\bm{Q}^+\bm{F}$, with $\bm{Q}^\# \in \mathcal{G} \subset \mathrm{SO}(3)$ and $\bm{Q}^+ \in \mathrm{SO}(3)$ must not change the material's response.
The special orthogonal group is denoted by $\mathrm{SO}(3)$, while $\mathcal{G}$ is the material's symmetry group.
Furthermore, in order to account for growth and remodeling, we assume a multiplicative decomposition, $\bm{F}=\bm{F}_e\bm{F}_g$, into an elastic part, $\bm{F}_e$, and a growth-related part, $\bm{F}_g$ \cite{rodriguez1994}.
Unfortunately, we note an inherent rotational non-uniqueness of this decomposition, i.e., $\bm{F}=\bm{F}_e^*\bm{F}_g^*$ where $\bm{F}_e^* = \bm{F}_e\bm{Q}^{*^T}$ and $\bm{F}_g^* = \bm{Q}^*\bm{F}_g$ with $\bm{Q}^* \in \mathrm{SO}(3)$ is equivalently possible (see \cite{casey2017}).\newline

\textbf{Stretch tensors.} Since the elastic and growth-related parts of $\bm{F}$ posses their polar decompositions, $\bm{F}_e = \bm{R}_e\bm{U}_e$ and $\bm{F}_g=\bm{V}_g\bm{R}_g$ with $\bm{R}_e,\bm{R}_g \in \mathrm{SO}(3)$, we found the stretch parts of both, $\bm{U}_e$ and $\bm{V}_g$, to be independent of $\bm{Q}^\#$ as well as $\bm{Q}^+$.
Hence, they are well suited to serve as arguments for the Helmholtz free energy $\psi$.
Unfortunately, as $\bm{U}_e^*=\bm{Q}^*\bm{U}_e\bm{Q}^{*^T}$ and $\bm{V}_g^*=\bm{Q}^*\bm{V}_g\bm{Q}^{*^T}$ the Helmholtz free energy suffers from the rotational non-uniqueness, i.e., $\psi\left(\bm{U}_e,\bm{V}_g\right)\neq\psi\left(\bm{U}_e^*,\bm{V}_g^*\right)$.
To overcome this issue, we restrict the energy to be a \textit{scalar-valued isotropic function}.\newline

\textbf{Co-rotated intermediate configuration.} As the elastic right Cauchy-Green tensor, $\bm{C}_e:=\bm{U}_e^2$, and the growth-related left Cauchy-Green tensor, $\bm{B}_g:=\bm{V}_g^2$, solely depend on the stretch tensors, we can alternatively express the energy $\psi=\psi\left(\bm{C}_e,\bm{B}_g\right)$.
Although the energy itself is unique and independent of any of the above mentioned rotations, its rate is not, i.e., $\partial\psi/\partial\bm{C}_e \neq \partial\psi/\partial\bm{C}_e^*$ and $\partial\psi/\partial\bm{B}_g \neq \partial\psi/\partial\bm{B}_g^*$.
Therefore, we employ the concept of a co-rotated intermediate configuration \cite{holthusen2023} and introduce the co-rotated elastic right Cauchy-Green tensor
\begin{equation}
	\bar{\bm{C}}_e := \bm{U}_g^{-1}\bm{C}\bm{U}_g^{-1}
\label{eq:Cebar}
\end{equation} 
and its growth-related counterpart, $\bm{C}_g:=\bm{F}_g^T\bm{F}_g$, where $\bm{C}:=\bm{F}^T\bm{F}$ is the right Cauchy-Green tensor.
Since both $\bar{\bm{C}}_e$ and $\bm{C}_e$ as well as $\bm{C}_g$ and $\bm{B}_g$ are \textit{similar}, we can exchange the arguments of the energy $\psi=\psi\left(\bm{C}_e,\bm{B}_g\right)=\psi\left(\bar{\bm{C}}_e,\bm{C}_g\right)$.
Noteworthy, the rate of the energy with respect to the co-rotated quantities is unique, which is considered an advantage.\newline

\textbf{Thermodynamic considerations.} Our models must be chosen in accordance with thermodynamics, i.e., the Clausius-Planck inequality for open systems $-\dot{\psi}+1/2\,\bm{S}:\dot{\bm{C}} + \mathcal{S}_0 \geq 0$ (see \cite{kuhl2003}) must be satisfied.
In the latter, $\bm{S}$ denotes the second Piola-Kirchhoff stress tensor, while $\mathcal{S}_0$ accounts for entropy sources and sinks resulting from interactions with the exterior of the system, e.g., the inflow of nutrients and hormones.
For growth and remodeling, we assume the Helmholtz free energy, $\psi$, to solely depend on the elastic stretches, i.e., $\psi=\psi\left(\bar{\bm{C}}_e\right)$.
Hence, we obtain the following (cf. \cite{holthusen2023})
\begin{equation}
	\left(\bm{S} - 2\,\bm{U}_g^{-1}\frac{\partial\psi}{\partial\bar{\bm{C}}_e}\bm{U}_g^{-1} \right) : \frac{1}{2}\dot{\bm{C}} + \underbrace{2\,\bar{\bm{C}}_e\frac{\partial\psi}{\partial\bar{\bm{C}}_e}}_{=: \bar{\bm{\Sigma}}} : \underbrace{\dot{\bm{U}}_g\,\bm{U}_g^{-1}}_{=: \bar{\bm{L}}_g} + \mathcal{S}_0 \geq 0
\label{eq:dissipation}
\end{equation}
from which we obtain the state law $\bm{S}=2\,\bm{U}_g^{-1}(\partial\psi/\partial\bar{\bm{C}}_e)\bm{U}_g^{-1}$ due to the arguments of \cite{coleman1961,coleman1963,coleman1967}.
As mentioned above, $\psi$ is a scalar-valued isotropic function of $\bar{\bm{C}}_e$, and thus, the driving force, $\bar{\bm{\Sigma}}$, is symmetric (cf. \cite{svendsen2001}).
Therefore, we can reduce the dissipation inequality 
\begin{equation}
	\bar{\bm{\Sigma}} : \bar{\bm{D}}_g + \mathcal{S}_0 \geq 0	
\end{equation}
with $\bar{\bm{D}}_g$ being the symmetric part of $\bar{\bm{L}}_g$.\newline
We must satisfy the reduced dissipation inequality at any time, and therefore, choose an appropriate evolution equation.
To this end, we follow the approach of \citet{lamm2022} and take inspiration from the well-established concept of pseudo potentials to model other inelastic phenomena such as elasto-plasticity, visco-elasticity, or phase transformations.
In accordance with \citet{kerstin1969}, we postulate the existence of a \textit{convex}, \textit{zero-valued}, and \textit{non-negative} pseudo potential, $g$, which solely depends on $\bar{\bm{\Sigma}}$.
Hence, our evolution equation reads
\begin{equation}
	\bar{\bm{D}}_g = \gamma\,\frac{\partial g}{\partial \bar{\bm{\Sigma}}}
\label{eq:Dg}
\end{equation}
where $\gamma$ denotes the growth multiplier.
Since $g$ is convex, the following identity is satisfied
\begin{equation}
	g\left(\bm{A}\right) \geq g\left(\bm{B}\right) + \frac{\partial g}{\partial \bm{B}} : \left(\bm{A} - \bm{B} \right) \quad \forall \bm{A},\bm{B} \in \mathbb{R}^{3\times 3}.
\end{equation}
For the following, we set $\bm{A}=\bm{0}$ and $\bm{B}=\bar{\bm{\Sigma}}$.
Furthermore, since $g\left(\bm{0}\right)=0$ as well as $g$ is non-negative, we obtain the following
\begin{equation}
	\frac{\partial g}{\partial \bar{\bm{\Sigma}}} : \bar{\bm{\Sigma}} \geq g\left(\bar{\bm{\Sigma}}\right) \geq 0.
\end{equation}
Thus, we recognize that the sign of the dissipation caused by $\bar{\bm{\Sigma}} : \bar{\bm{D}}_g$ is the same as the sign of the growth multiplier itself.\footnote{Noteworthy, for elasto-plasticity and visco-elasticity the sign of the multiplier is always greater or equal to zero. Since living tissues are open systems, the multiplier might be negative in case of growth and remodeling.}

\subsection{Helmholtz free energy}
\label{sec:Helmholtz}
As discussed in the previous section, stresses are caused by strains due to the existence of a Helmholtz free energy.
In the following, we investigate physically plausible restrictions of the energy and choose a general functional form, which will later on serve to design our architecture-based neural network to discover the material's behavior.\newline

\textbf{Restrictions.} Our energy must be chosen in such a way that self-penetration of the material is prevented.
In addition, increasing the deformation should result in an increase of the energy as well.
Hence, our energy must meet the conditions
\begin{equation}
	\lim_{\mathrm{det}\left(\bar{\bm{C}}_e\right) \to \infty} \psi \to +\infty \quad \land \lim_{\mathrm{det}\left(\bar{\bm{C}}_e\right) \searrow 0} \psi \to +\infty.
\end{equation}
which is referred to as the volumetric growth condition.
Furthermore, the normalization of the (elastic) Helmholtz free energy as well as the stress must be satisfied, i.e. for $\bar{\bm{C}}_e=\bm{I}$, both $\psi$ and $\bm{S}$ must be zero.
In addition, to guarantee the existence of at least one minimizing deformation, we design our energy to satisfy polyconvexity (see \cite{ball1976}).
For a more comprehensive overview, we refer the interested reader to \citet{kissas2024} and the literature cited therein.\newline

\textbf{Design of energy.} For the design of the energy, we employ the multiplicative decomposition of $\bar{\bm{C}}_e$ into its isochoric part, $\mathrm{det}\left(\bar{\bm{C}}_e\right)^{-1/3}\bar{\bm{C}}_e$, and its volumetric part, $\mathrm{det}\left(\bar{\bm{C}}_e\right)^{1/3}\bm{I}$, (cf. \cite{flory1961}) and split the energy additively into isochoric and volumetric contributions, i.e.,
\begin{equation}
\psi = \psi_{iso}\left(\mathrm{det}\left(\bar{\bm{C}}_e\right)^{-1/3}\bar{\bm{C}}_e\right) + \psi_{vol}\left(\mathrm{det}\left(\bar{\bm{C}}_e\right)\right).
\end{equation}
According to \citet{hartmann2003}, the sum of polyconvex functions is polyconvex.
For the volumetric part, $\psi_{vol}$, polyconvexity is guaranteed if the function is convex with respect to the determinant.
We choose a generic function suggested by \cite{ogden1972com} for this part
\begin{equation}
	\psi_{vol} = \mathrm{det}\left(\bar{\bm{C}}_e\right)^{-\beta} - 1 + \beta\,\ln\left(\mathrm{det}\left(\bar{\bm{C}}_e\right)\right).
\label{eq:Energy_vol}
\end{equation}
Since the second derivative of $\psi_{vol}$ is greater or equal to zero if $\beta\leq 0$, the volumetric part of the energy is convex.
It remains to choose a functional form for the isochoric part, $\psi_{iso}$, of the energy.
For living tissues, the Ogden model \cite{ogden1972incom} is well-established and polyconvex (see \cite{hartmann2003})
\begin{equation}
	\psi_{iso} = \sum_{i=1}^p \sum_{j=1}^3 \left(\mu_j^{\alpha_i} - 1 \right)
\label{eq:Energy_iso}
\end{equation}
where $\mu_j=\mathrm{det}\left(\bar{\bm{C}}_e\right)^{-1/3}\lambda_j$ are the eigenvalues of the isochoric part of $\bar{\bm{C}}_e=\sum_{i=1}^3\lambda_i\, \bm{m}_i \otimes \bm{m}_i$ with $\bm{m}_i$ denoting the corresponding eigenvectors (see \cite{steinmann2012}).
Both the number of $p$-Ogden terms and Ogden-exponents, $\alpha_i$, can be adjusted according to the data.
%
\subsection{Concept of pseudo potential and homeostatic surface}
\label{sec:potential}
In this contribution, we employ the concept of a homeostatic surface, $\Phi$, introduced by \citet{lamm2022}.
Similar to elasto-plasticity, homeostatic surfaces are defined in the principal stress space of the thermodynamically consistent driving force.
They describe the homeostasis of living tissues, and thus, the tissue grows and remodels in order to achieve a stress state that lies on this very surface.
This behavior is fundamentally different to elasto-plasticity, since the interior of the surface cannot be interpreted as an elastic regime.\newline

\textbf{Rate-independent case.} In order to make the idea of homeostatic surfaces clear, we begin with the (theoretical) case of rate-independent growth and remodeling.
In this case, the tissue must satisfy at any time that
\begin{equation}
	\Phi\left(\bar{\bm{\Sigma}}\right) := \phi\left(\bar{\bm{\Sigma}}\right) - \sigma_{hom} = 0
\label{eq:homeostatic_surface}
\end{equation}
where $\sigma_{hom}$ denotes the value of the homeostatic stress, while $\phi$ is a convex, zero-valued, and non-negative potential.
In analogy to elasto-plasticity, we call the case of $\phi\equiv g$ associative growth.
In this contribution, we restrict ourselves to this case.
Additionally, we restrict both $\phi$ and $g$ to be scalar-valued isotropic functions.
In order to satisfy Equation~\eqref{eq:homeostatic_surface} at any time, the growth multiplier, $\gamma$, must be determined accordingly.\newline

\textbf{Rate-dependent case.} Growth and remodeling is not an instantaneous process. To account for this time dependency, we employ a Perzyna-type approach (cf. \cite{perzyna1966,perzyna1971}).
Thus, we introduce the following
\begin{equation}
	\gamma = \frac{1}{\eta} \frac{\Phi}{\sigma_{hom}}
\end{equation}
with the `relaxation' or `growth and remodeling' time $\eta$.
Rewriting the latter equation yields
\begin{equation}
	\Phi - \gamma\,\eta\,\sigma_{hom} = 0
\label{eq:Perzyna}
\end{equation}
which must be satisfied at any time for the rate-dependent case.
Conceptually, we allow an under- and overstress of the homeostatic surface by subtracting $\gamma\,\eta\,\sigma_{hom}$.
Over time, both $\Phi$ and $\gamma$ tend to zero.
Noteworthy, Equation~\eqref{eq:Perzyna} reduces to Equation~\eqref{eq:homeostatic_surface} if $\eta=0$.\newline

\textbf{Design of potential.} It remains to choose a functional form of the homeostatic surface and the pseudo potential, respectively.
In line with e.g. \citet{soleimani2020}, it seems reasonable to express it in terms of principal stresses.
Here, we additively decompose the potential into a contribution of the principal stresses themselves as well as a potential depending on the shear stresses
\begin{equation}
	\phi\left(\bar{\bm{\Sigma}}\right) = \phi_\sigma(\sigma_1,\sigma_2,\sigma_3) + \phi_\tau(\tau_1,\tau_2,\tau_3)
\label{eq:potential_choice}
\end{equation}
where $\sigma_1 \geq \sigma_2 \geq \sigma_3$ denote the principal stresses of $\bar{\bm{\Sigma}}=\sum_{i=1}^3 \sigma_i\,\bm{n}_i\otimes\bm{n}_i$ with the corresponding eigenvectors $\bm{n}_i$.
In addition, the principal shear stresses are defined as
\begin{equation}
	\tau_1 = \frac{\sigma_1-\sigma_3}{2}; \quad \tau_2 = \frac{\sigma_1-\sigma_2}{2}; \quad \tau_3 = \frac{\sigma_2-\sigma_3}{2}.
\label{eq:principalshear}
\end{equation}
Since the principal stresses are ordered in a descending manner, we realize that the first principal shear stress, $\tau_1$, is the largest, and further, all shear stresses are greater or equal to zero.\newline
From a mechanical point of view, shear stresses are unlike to change the volume.
More precisely, we expect that a change in the determinant of $\bm{U}_g$ solely depends on the hydrostatic part of $\bar{\bm{\Sigma}}$, while isochoric deformations solely depend on its deviatoric part.
Therefore, we investigate the rates of the isochoric invariants \cite{steinmann2012}, $\tilde{I}_1^{\bm{U}_g} := \mathrm{tr}\left(\bm{U}_g\right)/\mathrm{det}\left(\bm{U}_g\right)^{1/3}$ and $\tilde{I}_2^{\bm{U}_g} := 1/2\left(\mathrm{tr}\left(\bm{U}_g\right)^2-\mathrm{tr}\left(\bm{U}_g^2\right)\right)/\mathrm{det}\left(\bm{U}_g\right)^{2/3}$, and the determinant
\begin{align}
	\frac{\mathrm{d}\tilde{I}_1^{\bm{U}_g}}{\mathrm{d}t} &= \mathrm{det}\left(\bm{U}_g\right)^{-1/3} \mathrm{dev}\left(\bm{U}_g\right) : \mathrm{dev}\left(\bar{\bm{D}}_g\right) \label{eq:dotI1_Ug}\\
	\frac{\mathrm{d}\tilde{I}_2^{\bm{U}_g}}{\mathrm{d}t} &= \mathrm{det}\left(\bm{U}_g\right)^{-2/3} \mathrm{dev}\left(\mathrm{tr}\left(\bm{U}_g\right)\bm{U}_g-\bm{U}_g^2\right) : \mathrm{dev}\left(\bar{\bm{D}}_g\right) \\
	\frac{\mathrm{d}\,\mathrm{det}\left(\bm{U}_g\right)}{\mathrm{d}t} &= \mathrm{det}\left(\bm{U}_g\right)\, \mathrm{tr}\left(\bar{\bm{D}}_g\right).
\end{align}
Further, we insert our choice~\eqref{eq:potential_choice} into Equation~\eqref{eq:Dg} and end up with the following
\begin{equation}
	\bar{\bm{D}}_g = \gamma\left(\frac{\partial\phi_\sigma}{\partial\bar{\bm{\Sigma}}} + \frac{\partial\phi_\tau}{\partial\bar{\bm{\Sigma}}} \right).
\end{equation}
Lastly, by evaluating the derivatives of the sub-potentials
\begin{align}
	\frac{\partial\phi_\sigma}{\partial\bar{\bm{\Sigma}}}  &= \sum_{i=1}^3 \frac{\partial\phi_\sigma}{\partial\sigma_i} \bm{n}_i \otimes \bm{n}_i \\
	2\,\frac{\partial\phi_\tau}{\partial\bar{\bm{\Sigma}}}  &= \frac{\partial\phi_\tau}{\partial\tau_1} \left( \bm{n}_1 \otimes \bm{n}_1 - \bm{n}_3 \otimes \bm{n}_3 \right) + \frac{\partial\phi_\tau}{\partial\tau_2} \left( \bm{n}_1 \otimes \bm{n}_1 - \bm{n}_2 \otimes \bm{n}_2 \right) + \frac{\partial\phi_\tau}{\partial\tau_3} \left( \bm{n}_2 \otimes \bm{n}_2 - \bm{n}_3 \otimes \bm{n}_3 \right) \label{eq:derivPhitau}
\end{align}
we observe that the sub-potential $\phi_\tau$ does not affect the rate of the determinant, since the trace of Equation~\eqref{eq:derivPhitau} is equal to zero, and thus, Equation~\eqref{eq:derivPhitau} is equivalent to its deviatoric part.
\section{Neural network}
\label{sec:network}
\textbf{Architecture.} The entire network consists of two feed-forward networks, one for the Helmholtz free energy and one for the pseudo potential, embedded into a recurrent network architecture to take time dependencies into account.
These feed-forward networks have a similar structure, i.e. the first layer calculates the eigenvalues of the tensorial input, while the subsequent layer applies custom-designed activation functions to ensure e.g. polyconvexity of the energy and convexity of the pseudo potential.
Generally, the weights of the networks can be classified into two types.
The first type influences the shape of the corresponding curve while the second scale the contribution of each neuron.
These last weights have a clear physical interpretation, for example, the shear and bulk modulus.
A distinguish feature of the iCANN for growth and remodeling in contrast to visco-elasticity is the necessity to satisfy the homeostatic surface equation.
To this end, we use a local Netwon-Raphson iteration.
We implemented the neural network into the open-source library \textit{TensorFlow} using the open-source interface \textit{Keras}.
The loss during training is chosen as the mean squared error between the experimentally obtained data and the stress predicted by our neural network.\newline

\textbf{Time discretization.} To solve Equation~\eqref{eq:Dg}, we need to employ a suitable time discretization scheme.
Therefore, we define the time interval $t \in \left[t_n,t_{n+1}\right]$.
An explicit exponential integrator scheme is used to keep the numerical effort during training the network low, but also to ensure that only the sub-potential $\phi_\sigma$ influences the determinant of $\bm{U}_g$.
For a more detailed explanation, we kindly refer to \cite{holthusen2024}.
With $\dot{\bm{C}}_g=2\,\bm{U}_g\bar{\bm{D}}_g\bm{U}_g^{-1}\bm{C}_g$ at hand, our update scheme reads as follows
\begin{equation}
	\bm{C}_{g_{n+1}} = \bm{U}_{g_n}\ \mathrm{exp}\left( 2\,\Delta t\, \gamma\, \frac{\partial \phi\left(\bar{\bm{\Sigma}}_n\right)}{\partial\bar{\bm{\Sigma}}_n} \right)\ \bm{U}_{g_n}, \quad \bm{U}_{g_{n+1}} = +\sqrt{\bm{C}_{g_{n+1}}}
\label{eq:time_discretization}
\end{equation}
with $\Delta t := t_{n+1} - t_n$.
Noteworthy, the latter cannot be directly solved, since the growth multiplier must be determined using a local Newton-Raphson iteration.\newline

\textbf{Newton-Raphson iteration.} In each time step, we have to meet the rate-dependent homeostatic surface Equation~\eqref{eq:Perzyna}.
To this end, we employ a local Newton-Raphson iteration within the architecture of our network.
Having in mind Equations~\eqref{eq:Cebar}, \eqref{eq:dissipation}, \eqref{eq:homeostatic_surface}, and \eqref{eq:time_discretization}, we can deduce that Equation~\eqref{eq:Perzyna} is solely a function of the unknown $\gamma$ since $\Phi_{n+1} = \Phi\left(\bar{\bm{\Sigma}}_{n+1}\left(\bar{\bm{C}}_{e_{n+1}}\left(\bm{C}_{n+1},\bm{U}_{g_{n+1}}\left(\gamma\right)\right)\right)\right)$.
To perform the Newton-Raphson iteration, we compute the jacobian using algorithmic differentiation.
For a better understanding of the iteration procedure, Algorithm~\ref{algo:Newton} provides a pseudo code implemented within the network's architecture.
For reasons which will be discussed in Section~\ref{sec:FFN_potential}, we replace some quantities in the code by their scaled version, $\hat{(\bullet)}$.
\begin{algorithm}[h]
\caption{Newton-Raphson iteration to determine $\hat{\gamma}$}
\label{algo:Newton}
\begin{algorithmic}[1]
\State \textbf{Input:} tolerance \( \epsilon \), weights $\mathbf{w}$, $\Delta t$, $\bm{C}_{n+1}$, $\frac{\partial \hat{\phi}\left(\bar{\bm{\Sigma}}_n\right)}{\partial\bar{\bm{\Sigma}}_n}$, $\bm{U}_{g_n}$
\State \textbf{Output:} $\hat{\gamma}$
\State Set \( \hat{\gamma} \gets 0 \)
\For{$k = 1$ to $30$}
	\State $\bm{C}_{g_{n+1}} \gets \bm{U}_{g_n}\ \mathrm{exp}\left( 2\,\Delta t\, \hat{\gamma}\, \frac{\partial \hat{\phi}\left(\bar{\bm{\Sigma}}_n\right)}{\partial\bar{\bm{\Sigma}}_n} \right)\ \bm{U}_{g_n}$ \Comment{cf. Equation~\eqref{eq:Dg_hat}}
	\State $\bm{U}_{g_{n+1}} \gets +\sqrt{\bm{C}_{g_{n+1}}}$
	\State $\bar{\bm{C}}_{e_{n+1}} \gets \bm{U}_{g_{n+1}}^{-1}\bm{C}_{n+1}\bm{U}_{g_{n+1}}^{-1}$
    \State Compute $\psi\left(\bar{\bm{C}}_{e_{n+1}}\right)$ and $\bar{\bm{\Sigma}}_{n+1}$
    \State Compute $\hat{\phi}\left(\bar{\bm{\Sigma}}_{{n+1}}\right)$
    \State $\hat{r} \gets \hat{\phi}\left(\bar{\bm{\Sigma}}_{{n+1}}\right) - 1 - \hat{\gamma}\,\hat{w}_\eta$ \Comment{see Equation~\eqref{eq:residual_hat}}
    \If{$|\hat{r}| < \epsilon$}
        \State \textbf{break}
    \EndIf    
    \State $\hat{r}' \gets \frac{\partial \hat{r}}{\partial \hat{\gamma}}$
    \State $\hat{\gamma} \gets \hat{\gamma} - \frac{\hat{r}}{\hat{r}'}$
\EndFor
\State \textbf{return} $\gamma$
\end{algorithmic}
\end{algorithm}
%
\subsection{Design of feed-forward networks}
\label{sec:FFN}
The presented constitutive framework and network architecture are broadly applicable. To capture the specific behavior of living tissues, we must design governing scalar quantities like energy and potential. This contribution focuses on how to achieve such designs while incorporating constitutive knowledge upfront. The physics-based architecture significantly reduces the number of weights in (inelastic) CANNs compared to standard neural networks. This reduction improves training robustness but may limit network generality — both perspectives have merit. Our goal is to outline the fundamental design strategy, reducing each network to a minimal size, though complexity can be easily scaled, as demonstrated in \cite{stpierre2023} and \cite{holthusen2024,holthusen2024PAMM}.
%
\subsubsection{Feed-forward network: Helmholtz free energy}
\label{sec:FFN_energy}
\begin{figure}[h]
	\centering
	\begin{tikzpicture}

\node[draw,rounded corners=0.1cm,minimum width=1.5cm,fill=white] (C) at (-8,0) {$\bar{\bm{C}}_e$};

\node[draw,rounded corners=0.1cm,minimum width=1.5cm, fill=rwthb4] (psi) at (7.5,0) {$\psi(\bar{\bm{C}}_e)$};


\node[draw,rounded corners=0.1cm,minimum width=2cm,fill=rwthb3] (vol) at (2.1,-3.6) {$(\bullet)^{w_{0,1}^\psi}-1-w_{0,1}^\psi\,\mathrm{ln}(\bullet)$};


\node[draw,rounded corners=0.1cm,minimum width=2cm,fill=rwthb3] (iso1) at (1,3.6) {$(\bullet)^{w_{1,1}^\psi}-1$};

\node[draw,rounded corners=0.1cm,minimum width=2cm,fill=rwthb3] (iso2) at (1,1.2) {$(\bullet)^{w_{1,1}^\psi}-1$};

\node[draw,rounded corners=0.1cm,minimum width=2cm,fill=rwthb3] (iso3) at (1,-1.2) {$(\bullet)^{w_{1,1}^\psi}-1$};

\node[draw,rounded corners=0.1cm,minimum width=2cm,fill=rwthb3] (iso4) at (4,1.2) {$(\bullet)$};


\node[draw,rounded corners=0.1cm,minimum width=2cm,fill=rwthb1] (lam1) at (-5.5,2.5) {$\lambda_1$};

\node[draw,rounded corners=0.1cm,minimum width=2cm,fill=rwthb1] (lam2) at (-5.5,0) {$\lambda_2$};

\node[draw,rounded corners=0.1cm,minimum width=2cm,fill=rwthb1] (lam3) at (-5.5,-2.5) {$\lambda_3$};

\node[draw,rounded corners=0.1cm,minimum width=2cm,fill=rwthb2] (mu1) at (-2.5,3.6) {$\mu_1$};

\node[draw,rounded corners=0.1cm,minimum width=2cm,fill=rwthb2] (mu2) at (-2.5,1.2) {$\mu_2$};

\node[draw,rounded corners=0.1cm,minimum width=2cm,fill=rwthb2] (mu3) at (-2.5,-1.2) {$\mu_3$};

\node[draw,rounded corners=0.1cm,minimum width=2cm,fill=rwthb2] (det) at (-2.5,-3.6) {$\mathrm{det}\left(\bar{\bm{C}}_e\right)$};


\draw[->] (C.east) to (lam1.west);

\draw[->] (C.east) to (lam2.west);

\draw[->] (C.east) to (lam3.west);

\draw[->] (lam1.east) to (mu1.west);

\draw[->] (lam1.east) to (mu2.west);

\draw[->] (lam1.east) to (mu3.west);

\draw[->] (lam1.east) to (det.west);

\draw[->] (lam2.east) to (mu1.west);

\draw[->] (lam2.east) to (mu2.west);

\draw[->] (lam2.east) to (mu3.west);

\draw[->] (lam2.east) to (det.west);

\draw[->] (lam3.east) to (mu1.west);

\draw[->] (lam3.east) to (mu2.west);

\draw[->] (lam3.east) to (mu3.west);

\draw[->] (lam3.east) to (det.west);

\draw[->] (mu1.east) to (iso1.west);

\draw[->] (mu2.east) to (iso2.west);

\draw[->] (mu3.east) to (iso3.west);

\draw[->] (det.east) to (vol.west);

\draw[->] (iso1.east) to (iso4.west);

\draw[->] (iso2.east) to (iso4.west);

\draw[->] (iso3.east) to (iso4.west);

\draw[->] (iso4.east) to (psi.west);

\draw[->] (vol.east) to node[above, yshift=5pt, xshift=-5pt] {$w_{0,2}^{\psi}$} (psi.west);

\draw[->] (iso4.east) to node[above, yshift=5pt] {$w_{1,2}^{\psi}$} (psi.west);



\end{tikzpicture} 
	\caption{Schematic illustration of our feed-forward network for the Helmholtz free energy, $\psi$, embedded into the recurrent network architecture. The first layer computes the eigenvalues of $\bar{\bm{C}}_e$, subsequently the deformation is decomposed into isochoric and volumetric deformations. The last layer applies custom-designed activation functions including a first set of weights, $w_{\star,1}^\psi$. These functions are multiplied by a second set of weights, $w_{\star,2}^\psi$. Noteworthy, $\mathrm{det}\left(\bar{\bm{C}}_e\right)=\lambda_1\lambda_2\lambda_3$ holds.}
\label{fig:FFN_energy}
\end{figure}
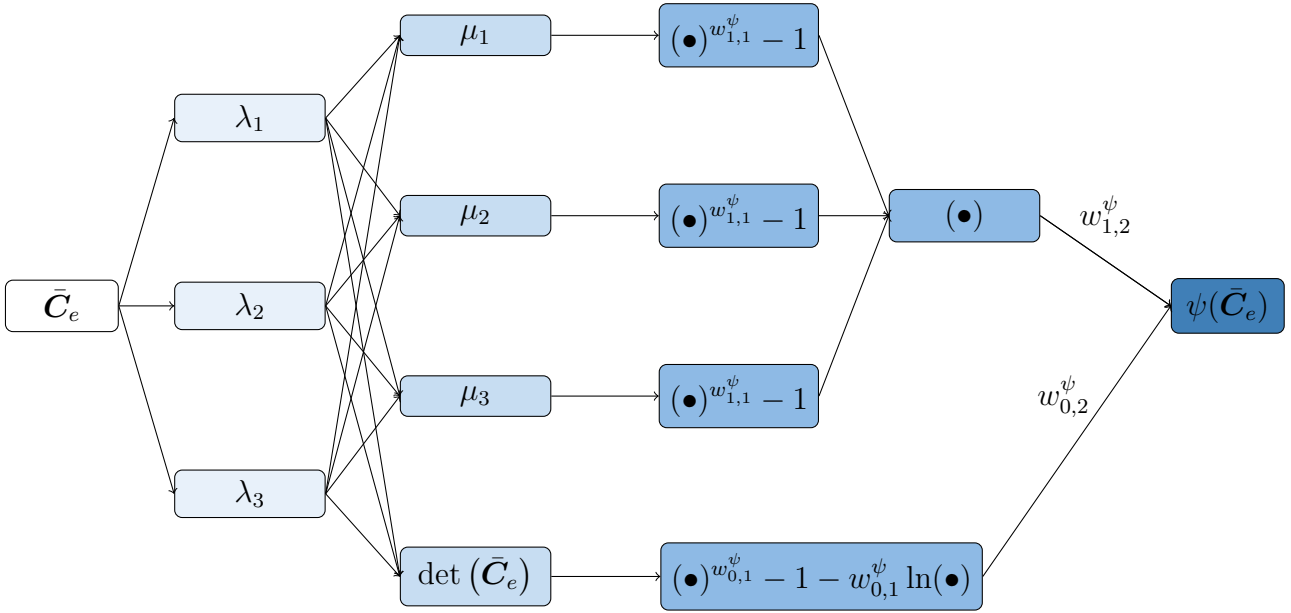

As mentioned in Section~\ref{sec:Helmholtz}, we decompose the energy according to the isochoric-volumetric decomposition of $\bar{\bm{C}}_e$ into an isochoric and a volumetric contribution, i.e. $\psi=\psi_{iso}\left(\mu_1,\mu_2,\mu_3\right) + \psi_{vol}\left(\mathrm{det}\left(\bar{\bm{C}}_e\right)\right)$.
Thus, it is straightforward to ensure polyconvexity, the volumetric growth condition, and the normalization of the energy.
The feed-forward network's architecture of the energy is illustrated in Figure~\ref{fig:FFN_energy} and can be written as follows
\begin{equation}
\begin{aligned}
	\psi &= w_{0,2}^\psi \left( \mathrm{det}\left(\bar{\bm{C}}_e\right)^{w_{0,1}^\psi} - 1 - w_{0,1}^\psi\ \mathrm{ln}\left(\mathrm{det}\left(\bar{\bm{C}}_e\right)\right)\right) \\
	&+ w_{1,2}^\psi \left( \mu_1^{w_{1,1}^\psi} + \mu_2^{w_{1,1}^\psi} + \mu_3^{w_{1,1}^\psi} - 3 \right).
\end{aligned}
\label{eq:ffn_energy}
\end{equation}
In this sense, $w_{0,2}^\psi$ can be mechanically interpreted as the material's elastic bulk modulus, while the elastic shear modulus is denoted by $w_{1,2}^\psi$.
Further, $w_{0,1}^\psi$ must be greater than or equal to zero to guarantee convexity of the volumetric contribution.
Of course, we can increase the complexity of both energy contributions arbitrarily.
For instance, the number of $p$-Ogden terms in Equation~\eqref{eq:Energy_iso} can be increased which increases the number of neurons in our network, respectively.
Similar, we might add polynomial terms to the volumetric energy (see \cite{holthusen2024}).
However, for the purpose of this contribution, our choices are sufficient.
%
\subsubsection{Feed-forward network: Pseudo potential}
\label{sec:FFN_potential}
Since we consider associative growth in this contribution, our network discovers only the potential $\phi$.
Further, we need to calculate the growth multiplier, $\gamma$, in each iteration step to satisfy Equation~\eqref{eq:Perzyna}.
Within the neural network, the parameters $\sigma_{hom}$ as well as $\eta$ are considered as weights to be determined during training.
However, we note that one of these parameters is redundant.
Therefore, we introduce the modified homeostatic surface, $\hat{\Phi} := \Phi/\sigma_{hom} = \hat{\phi}\left(\bar{\bm{\Sigma}}\right) - 1$ with $\hat{\phi}=\phi/\sigma_{hom}$.
Thus, we solve the following equation within each step
\begin{equation}
	r(\gamma) := \hat{\phi} - 1 - \gamma\, w_\eta = 0
\label{eq:residual}
\end{equation}
where $w_\eta$ is a weight to be determined during training.
Due to the assumption of associative growth, we can exchange $g$ by $\phi$ in Equation~\eqref{eq:Dg}.
However, this means we have to discover both $\phi$ and $\hat{\phi}$.
Therefore, we adjust Equation~\eqref{eq:residual}, i.e.,
\begin{equation}
	\hat{r}(\hat{\gamma}) := \hat{\phi} - 1 - \hat{\gamma}\, \hat{w}_\eta = 0
\label{eq:residual_hat}
\end{equation}
where $\hat{\gamma} := \gamma\,\sigma_{hom}$ and $\hat{w}_\eta := \frac{w_\eta}{\sigma_{hom}}$.
Consequently, the evolution equation for $\bar{\bm{D}}_g$ can be expressed as follows
\begin{equation}
	\bar{\bm{D}}_g = \hat{\gamma}\,\frac{\partial\hat{\phi}}{\partial\bar{\bm{\Sigma}}}
\label{eq:Dg_hat}
\end{equation}
which we use in combination with Equation~\eqref{eq:residual_hat} within the network's architecture.
Noteworthy, this is only due to reduce the number of redundant weights.
From a mechanical point of view, the theoretical framework remains unchanged.

Next, we need to design a generic formalism of our network for the pseudo potential, which must be able to discover homeostasis.
Homeostasis is characterized by a profound asymmetry in tension and compression.
In particular, tensional homeostasis refers to a homeostatic state which is achieved due to the rise of tensional stresses.
To enable our network to differentiate between tension and compression, we propose a particular set of custom-designed activation functions: $\mathrm{max}\left(-(\bullet),0\right)$, $\mathrm{ln}\left(\mathrm{cosh}(\bullet)\right)$, $\mathrm{max}\left((\bullet),0\right)$.
The first activation function accounts only for compressive stresses, the second is symmetric with respect to tension and compression, and the last takes tensional stresses into account.
With these custom-designed functions at hand as well as the conceptual design discussed in Section~\ref{sec:potential}, we design our network as shown in Figure~\ref{fig:FFN_potenial}.
Equivalently, we can express the network in its equation form as
\begin{equation}
\begin{aligned}
	\hat{\phi} &=  w_{\sigma,1}^\phi \left( \mathrm{max}\left(-\sigma_1,0\right) + \mathrm{max}\left(-\sigma_2,0\right) + \mathrm{max}\left(-\sigma_3,0\right) \right) \\
		&+ w_{\sigma,2}^\phi \left( \mathrm{ln}\left(\mathrm{cosh}\left(w_{\sigma,3}^\phi\,\sigma_1\right)\right) + \mathrm{ln}\left(\mathrm{cosh}\left(w_{\sigma,3}^\phi\,\sigma_2\right)\right) + \mathrm{ln}\left(\mathrm{cosh}\left(w_{\sigma,3}^\phi\,\sigma_3\right)\right)\right) \\
		&+ w_{\sigma,4}^\phi \left( \mathrm{max}\left(\sigma_1,0\right) + \mathrm{max}\left(\sigma_2,0\right) + \mathrm{max}\left(\sigma_3,0\right) \right) \\
		&+ w_{\tau,1}^\phi \left( \mathrm{max}\left(-\tau_1,0\right) + \mathrm{max}\left(-\tau_2,0\right) + \mathrm{max}\left(-\tau_3,0\right) \right) \\
		&+ w_{\tau,2}^\phi \left( \mathrm{ln}\left(\mathrm{cosh}\left(w_{\tau,3}^\phi\,\tau_1\right)\right) + \mathrm{ln}\left(\mathrm{cosh}\left(w_{\tau,3}^\phi\,\tau_2\right)\right) + \mathrm{ln}\left(\mathrm{cosh}\left(w_{\tau,3}^\phi\,\tau_3\right)\right)\right) \\
		&+ w_{\tau,4}^\phi \left( \mathrm{max}\left(\tau_1,0\right) + \mathrm{max}\left(\tau_2,0\right) + \mathrm{max}\left(\tau_3,0\right) \right).
\end{aligned}
\label{eq:ffn_potential}
\end{equation}
Here, the weights $w_{\star,3}^\phi$ have a similar meaning as the weights $w_{\star,1}^\psi$ in case of the energy (cf. Section~\ref{sec:FFN_energy}).
The remaining weights tell us the value of the homeostatic stress, $\sigma_{hom}$, as well as the ratio between principal stresses and principal shear stresses.
Thus, we might also learn whether growth and remodeling takes place in an isochoric manner (cf. Equations~\eqref{eq:dotI1_Ug}-\eqref{eq:derivPhitau}).
\begin{figure}[!htp]
	\centering
	\begin{tikzpicture}

\node[draw,rounded corners=0.1cm,minimum width=1.5cm,fill=white] (S) at (-8,0) {$\bar{\bm{\Sigma}}$};

\node[draw,rounded corners=0.1cm,minimum width=1.5cm, fill=rwtho5] (phi) at (8.5,0) {$\hat{\phi}(\bar{\bm{\Sigma}})$};


\node[draw,rounded corners=0.1cm,minimum width=2cm,fill=rwtho3] (maxps1) at (1,7.3) {$\mathrm{max}\left(-(\bullet),0\right)$};

\node[draw,rounded corners=0.1cm,minimum width=2cm,fill=rwtho3] (lns1) at (1,6.5) {$\mathrm{ln}\left(\mathrm{cosh}\left(\bullet\right)\right)$};

\node[draw,rounded corners=0.1cm,minimum width=2cm,fill=rwtho3] (maxms1) at (1,5.7) {$\mathrm{max}\left((\bullet),0\right)$};

\node[draw,rounded corners=0.1cm,minimum width=2cm,fill=rwtho3] (maxps2) at (1,4.7) {$\mathrm{max}\left(-(\bullet),0\right)$};

\node[draw,rounded corners=0.1cm,minimum width=2cm,fill=rwtho3] (lns2) at (1,3.9) {$\mathrm{ln}\left(\mathrm{cosh}\left(\bullet\right)\right)$};

\node[draw,rounded corners=0.1cm,minimum width=2cm,fill=rwtho3] (maxms2) at (1,3.1) {$\mathrm{max}\left((\bullet),0\right)$};

\node[draw,rounded corners=0.1cm,minimum width=2cm,fill=rwtho3] (maxps3) at (1,2.1) {$\mathrm{max}\left(-(\bullet),0\right)$};

\node[draw,rounded corners=0.1cm,minimum width=2cm,fill=rwtho3] (lns3) at (1,1.3) {$\mathrm{ln}\left(\mathrm{cosh}\left(\bullet\right)\right)$};

\node[draw,rounded corners=0.1cm,minimum width=2cm,fill=rwtho3] (maxms3) at (1,0.5) {$\mathrm{max}\left((\bullet),0\right)$};

\node[draw,rounded corners=0.1cm,minimum width=2cm,fill=rwtho3] (maxpt1) at (1,-0.5) {$\mathrm{max}\left(-(\bullet),0\right)$};

\node[draw,rounded corners=0.1cm,minimum width=2cm,fill=rwtho3] (lnt1) at (1,-1.3) {$\mathrm{ln}\left(\mathrm{cosh}\left(\bullet\right)\right)$};

\node[draw,rounded corners=0.1cm,minimum width=2cm,fill=rwtho3] (maxmt1) at (1,-2.1) {$\mathrm{max}\left((\bullet),0\right)$};

\node[draw,rounded corners=0.1cm,minimum width=2cm,fill=rwtho3] (maxpt2) at (1,-3.1) {$\mathrm{max}\left(-(\bullet),0\right)$};

\node[draw,rounded corners=0.1cm,minimum width=2cm,fill=rwtho3] (lnt2) at (1,-3.9) {$\mathrm{ln}\left(\mathrm{cosh}\left(\bullet\right)\right)$};

\node[draw,rounded corners=0.1cm,minimum width=2cm,fill=rwtho3] (maxmt2) at (1,-4.7) {$\mathrm{max}\left((\bullet),0\right)$};

\node[draw,rounded corners=0.1cm,minimum width=2cm,fill=rwtho3] (maxpt3) at (1,-5.7) {$\mathrm{max}\left(-(\bullet),0\right)$};

\node[draw,rounded corners=0.1cm,minimum width=2cm,fill=rwtho3] (lnt3) at (1,-6.5) {$\mathrm{ln}\left(\mathrm{cosh}\left(\bullet\right)\right)$};

\node[draw,rounded corners=0.1cm,minimum width=2cm,fill=rwtho3] (maxmt3) at (1,-7.3) {$\mathrm{max}\left((\bullet),0\right)$};

\node[draw,rounded corners=0.1cm,minimum width=2cm,fill=rwtho4] (summaxms) at (4.5,6.5) {$\left(\bullet\right)$};

\node[draw,rounded corners=0.1cm,minimum width=2cm,fill=rwtho4] (sumlns) at (4.5,3.9) {$\left(\bullet\right)$};

\node[draw,rounded corners=0.1cm,minimum width=2cm,fill=rwtho4] (summaxps) at (4.5,1.3) {$\left(\bullet\right)$};

\node[draw,rounded corners=0.1cm,minimum width=2cm,fill=rwtho4] (summaxmt) at (4.5,-1.3) {$\left(\bullet\right)$};

\node[draw,rounded corners=0.1cm,minimum width=2cm,fill=rwtho4] (sumlnt) at (4.5,-3.9) {$\left(\bullet\right)$};

\node[draw,rounded corners=0.1cm,minimum width=2cm,fill=rwtho4] (summaxpt) at (4.5,-6.5) {$\left(\bullet\right)$};


\node[draw,rounded corners=0.1cm,minimum width=2cm,fill=rwtho1] (s1) at (-5.5,2.5) {$\sigma_1$};

\node[draw,rounded corners=0.1cm,minimum width=2cm,fill=rwtho1] (s2) at (-5.5,0) {$\sigma_2$};

\node[draw,rounded corners=0.1cm,minimum width=2cm,fill=rwtho1] (s3) at (-5.5,-2.5) {$\sigma_3$};

\node[draw,rounded corners=0.1cm,minimum width=2cm,fill=rwtho2] (s1o) at (-2.5,6.5) {$(\bullet)$};

\node[draw,rounded corners=0.1cm,minimum width=2cm,fill=rwtho2] (s2o) at (-2.5,3.9) {$(\bullet)$};

\node[draw,rounded corners=0.1cm,minimum width=2cm,fill=rwtho2] (s3o) at (-2.5,1.3) {$(\bullet)$};

\node[draw,rounded corners=0.1cm,minimum width=2cm,fill=rwtho2] (t1) at (-2.5,-1.3) {$\tau_1$};

\node[draw,rounded corners=0.1cm,minimum width=2cm,fill=rwtho2] (t2) at (-2.5,-3.9) {$\tau_2$};

\node[draw,rounded corners=0.1cm,minimum width=2cm,fill=rwtho2] (t3) at (-2.5,-6.5) {$\tau_3$};


\draw[->] (S.east) to (s1.west);

\draw[->] (S.east) to (s2.west);

\draw[->] (S.east) to (s3.west);

\draw[->] (s1.east) to (s1o.west);

\draw[->] (s1.east) to (t1.west);

\draw[->] (s1.east) to (t2.west);

\draw[->] (s2.east) to (s2o.west);

\draw[->] (s2.east) to (t2.west);

\draw[->] (s2.east) to (t3.west);

\draw[->] (s3.east) to (s3o.west);

\draw[->] (s3.east) to (t1.west);

\draw[->] (s3.east) to (t3.west);

\draw[->] (s1o.east) to (maxps1.west);

\draw[->] (s1o.east) to node[above, yshift=-4pt, xshift=8pt] {$w_{\sigma,3}^{\phi}$} (lns1.west);

\draw[->] (s1o.east) to (maxms1.west);

\draw[->] (s2o.east) to (maxps2.west);

\draw[->] (s2o.east) to node[above, yshift=-4pt, xshift=8pt] {$w_{\sigma,3}^{\phi}$}(lns2.west);

\draw[->] (s2o.east) to (maxms2.west);

\draw[->] (s3o.east) to (maxps3.west);

\draw[->] (s3o.east) to node[above, yshift=-4pt, xshift=8pt] {$w_{\sigma,3}^{\phi}$}(lns3.west);

\draw[->] (s3o.east) to (maxms3.west);

\draw[->] (t1.east) to (maxpt1.west);

\draw[->] (t1.east) to node[above, yshift=-4pt, xshift=8pt] {$w_{\tau,3}^{\phi}$}(lnt1.west);

\draw[->] (t1.east) to (maxmt1.west);

\draw[->] (t2.east) to (maxpt2.west);

\draw[->] (t2.east) to node[above, yshift=-4pt, xshift=8pt] {$w_{\tau,3}^{\phi}$}(lnt2.west);

\draw[->] (t2.east) to (maxmt2.west);

\draw[->] (t3.east) to (maxpt3.west);

\draw[->] (t3.east) to node[above, yshift=-4pt, xshift=8pt] {$w_{\tau,3}^{\phi}$}(lnt3.west);

\draw[->] (t3.east) to (maxmt3.west);

\draw[->] (maxps1.east) to (summaxms.west);

\draw[->] (maxps2.east) to (summaxms.west);

\draw[->] (maxps3.east) to (summaxms.west);

\draw[->] (lns1.east) to (sumlns.west);

\draw[->] (lns2.east) to (sumlns.west);

\draw[->] (lns3.east) to (sumlns.west);

\draw[->] (maxms1.east) to (summaxps.west);

\draw[->] (maxms2.east) to (summaxps.west);

\draw[->] (maxms3.east) to (summaxps.west);

\draw[->] (maxmt1.east) to (summaxpt.west);

\draw[->] (maxmt2.east) to (summaxpt.west);

\draw[->] (maxmt3.east) to (summaxpt.west);

\draw[->] (maxpt1.east) to (summaxmt.west);

\draw[->] (maxpt2.east) to (summaxmt.west);

\draw[->] (maxpt3.east) to (summaxmt.west);

\draw[->] (lnt1.east) to (sumlnt.west);

\draw[->] (lnt2.east) to (sumlnt.west);

\draw[->] (lnt3.east) to (sumlnt.west);

\draw[->] (maxmt1.east) to (summaxpt.west);

\draw[->] (maxmt2.east) to (summaxpt.west);

\draw[->] (maxmt3.east) to (summaxpt.west);

\draw[->] (summaxms.east) to node[above, yshift=5pt, xshift=8pt] {$w_{\sigma,1}^{\phi}$} (phi.west);

\draw[->] (sumlns.east) to node[above, yshift=0pt, xshift=-18pt] {$w_{\sigma,2}^{\phi}$} (phi.west);

\draw[->] (summaxps.east) to node[above, yshift=0pt, xshift=-1pt] {$w_{\sigma,4}^{\phi}$} (phi.west);

\draw[->] (summaxmt.east) to node[above, yshift=0pt, xshift=-1pt] {$w_{\tau,1}^{\phi}$} (phi.west);

\draw[->] (sumlnt.east) to node[above, yshift=-12pt, xshift=-18pt] {$w_{\tau,2}^{\phi}$} (phi.west);

\draw[->] (summaxpt.east) to node[above, yshift=-12pt, xshift=12pt] {$w_{\tau,4}^{\phi}$} (phi.west);

\end{tikzpicture} 
	\caption{Schematic illustration of our feed-forward network for the pseudo potential, $\hat{\phi}$, embedded into the recurrent network architecture. The first layer computes the eigenvalues of $\bar{\bm{\Sigma}}$, subsequently the principal shear stresses are computed. The last layer applies custom-designed activation functions including a first set of weights, $w_{\star,3}^\phi$. These functions are multiplied by a second set of weights, $w_{\star}^\phi$.}
	\label{fig:FFN_potenial}
\end{figure}
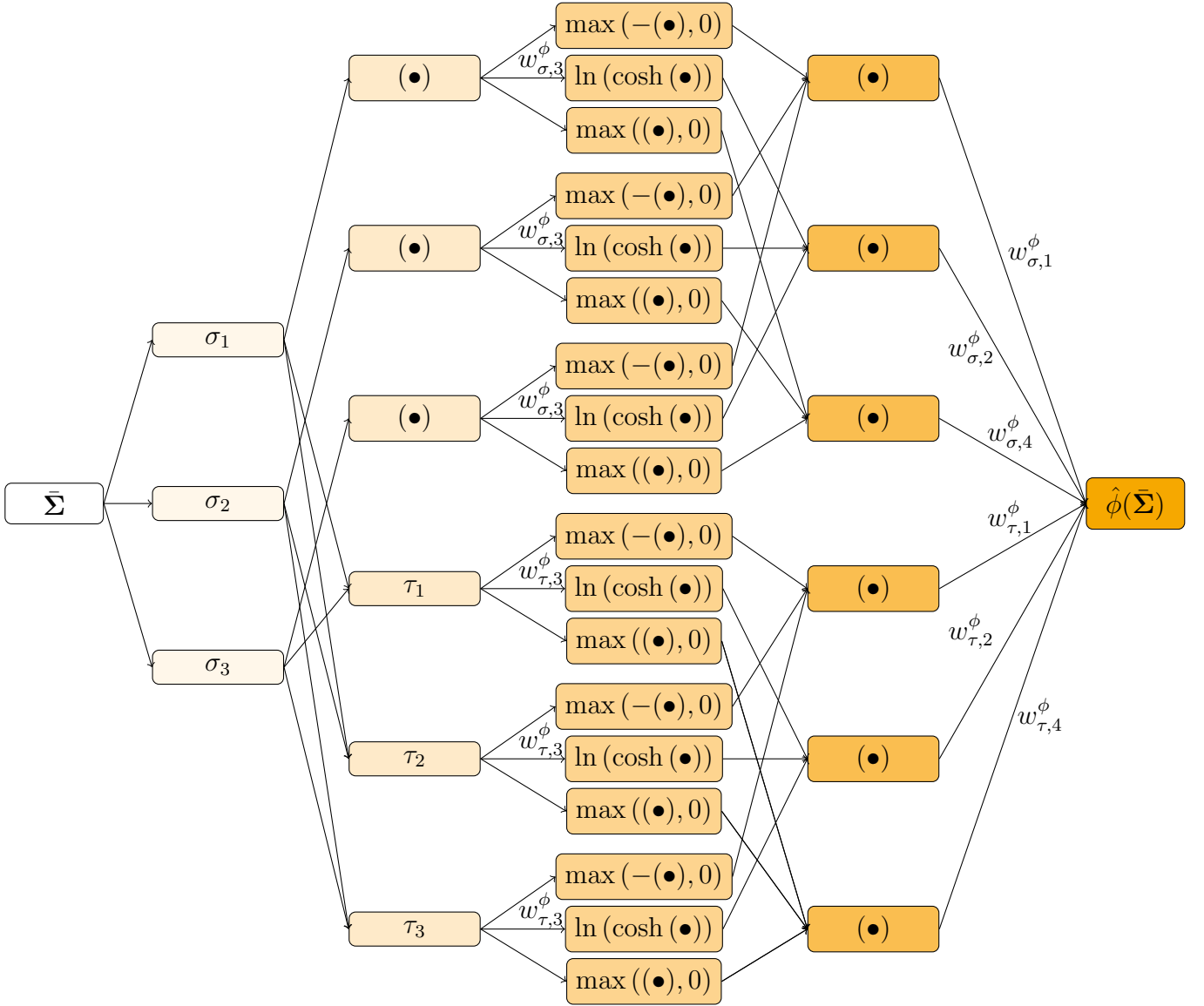
\begin{remark}
To discover \textit{tensional} homeostasis, we could exclude the activation function $\mathrm{max}\left(-(\bullet),0\right)$ a priori.
However, first to be able to discover possible compressional homeostasis, second to be more general, and third -- most importantly -- to keep user-biased discovery at a minimum, we include this particular function as well.
Further, due to the definition of the principal shear stresses~\eqref{eq:principalshear}, we could reduce the number of neurons in the network, since the shear stresses are greater or equal to zero.
Thus, $\mathrm{max}\left(-\tau_1,0\right) = \mathrm{max}\left(-\tau_2,0\right) = \mathrm{max}\left(-\tau_3,0\right) \equiv 0$ as well as $\mathrm{max}\left(\tau_1,0\right)=\tau_1$, $\mathrm{max}\left(\tau_2,0\right)=\tau_2$, and $\mathrm{max}\left(\tau_3,0\right)=\tau_3$.
Nevertheless, to illustrate the network in its most generic form, we do not further adjust the network.
Noteworthy, during training the corresponding weights should ideally turn out to be zero.
\end{remark}
\begin{remark}
Within the network architecture shown in Figure~\ref{fig:FFN_potenial}, we considered only one specific linear combination of principal stresses, namely the principal shear stresses.
Suppose we have a convex, zero-valued, and non-negative function, $f(g)$, with $g=a\,\sigma_1+b\,\sigma_2+c\,\sigma_3$, where $a,b,c \in \mathbb{R}$.
We can study the convexity of $f$ by evaluating the eigenvalues of its hessian.
In our case, the only non-zero eigenvalue of the hessian is $\frac{\partial^2 f}{\partial g^2}\,(a^2+b^2+c^2)$. 
Since $f(g)$ is convex, its second derivative is greater than or equal to zero, and thus the hessian is positive semidefinite.
Thus, $f$ is also convex with respect to the principal stresses.
This result allows us to linearly combine the principal stresses, and thus, to extend the architecture of the pseudo potential.
For example, it would be possible to consider the mean of the principal stresses, i.e., the hydrostatic pressure.
The works of e.g. \citet{lubarda2002} and \citet{himpel2005} describe growth and remodeling using the hydrostatic pressure.
More generally, we could consider $a,b,c$ as additional weights to be discovered during training.
For the time being, we consider the chosen architecture to be sufficient enough.
\end{remark}
\section{Results}
\label{sec:results}
In this section, we aim to explore the capabilities of iCANNs, enhanced by homeostatic surfaces, to automatically discover a material model for tissue equivalents, assess its predictive accuracy at the material point level, and evaluate its ability to qualitatively represent structural deformations. To this end, we investigate two experimental setups from the literature \cite{eichinger2020}.
Each setup follows two steps: (i) the specimen remains undeformed until tensile homeostasis is achieved after a period of time, and (ii) the homeostatic force is perturbed by either stretching or compressing the specimen to examine whether homeostasis can be restored.
The first setup involves a stripe specimen clamped at both edges (Section~\ref{sec:stripe}), while the second setup examines a cross specimen, where all four arms are clamped (Section~\ref{sec:cross}).
For a detailed description of the experimental setups, the preparation of the specimens and the testing device, we kindly refer to \cite{eichinger2020}.

Stress-strain data are required to train the neural network. 
From the force-time curves provided in the aforementioned work, we precomputed the engineering stress in the principal loading directions, known as the first Piola-Kirchhoff stress $\bm{F}\bm{S}$, by dividing the force by the cross-sectional area of $40\ \left[\text{mm}^2\right]$ of the stripe specimen and the arms of the cross specimen, respectively.
From this we calculated the second Piola-Kirchhoff stress.

In addition, training the network necessitates the entire stress and stretch tensors. 
Therefore, we assume that the stress state in the first experimental setup is nearly uniaxial. For the cross specimen, we similarly assume an approximately biaxial stress state. However, since the network's input is the right Cauchy Green tensor $\bm{C}$, we also need measurements of the full strain field, which are not provided in the experimental data.
Thus, we further assume that the stretch in the off-principal directions is negligible and that the corresponding stress is zero\footnote{In the first setup, this applies to two directions, while in the biaxial case, it applies only to the direction perpendicular to the biaxial loading.}. This approach is comparable to assuming an incompressible material behavior\footnote{Since no external load is applied in the first loading step, yet a rise in stress (homeostasis) is observed, assuming incompressibility leads to unphysical results.
The entire stress tensor $\bm{S}$ would be equal to the zero tensor.}.
We enforce these constraints by Lagrange multipliers, see \ref{app:lagrange} for the theoretical derivation. 

We acknowledge that the assumptions of homogeneous stress/strain states, negligible stretch, and zero associated stress affect our findings. However, at this stage of development, we consider these assumptions acceptable.

In all simulations, we used the \textit{MeanSquaredError} class provided by \textit{Keras}, which calculates the loss between the experimentally observed and predicted stress.
Further, for optimization we choose the \textit{ADAM} optimizer with a learning rate of $0.001$.
%
\subsection{Discovering a model for the stripe specimen}
\label{sec:stripe}
We begin with evaluating the iCANN for the stripe specimen.
The experimental curves, expressed in terms of the second Piola-Kirchhoff stress, are shown in Figure~\ref{fig:training_testing_stripe} for both stretching and compressing the specimen.
Homeostasis is reached after seventeen hours, during which no deformation is applied, i.e., $\bm{C}=\bm{I}$.
Subsequently, the force in the longitudinal direction is perturbed by $\pm 10\%$ of the homeostatic force measured at $t=17$ [h].
Noteworthy, this perturbation is conducted in a displacement-driven manner.
Consequently, the corresponding components of the right Cauchy Green tensor are $C_{11}=0.99505347$ [-] (compressing) and $C_{11}=1.0037114$ [-] (stretching).
As explained above, the remaining diagonal terms of $\bm{C}$ remain equal to one.\newline

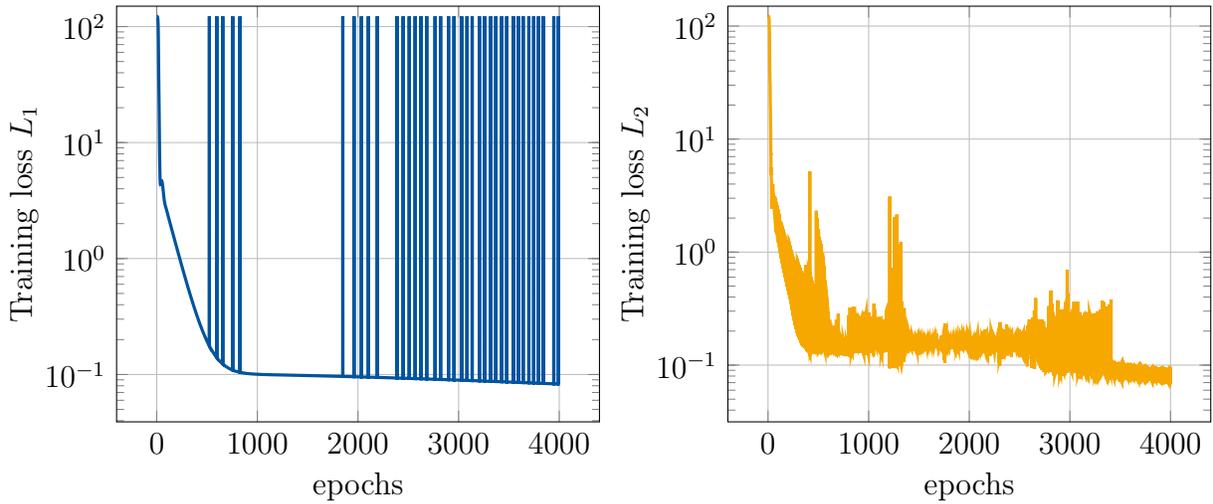
\begin{figure}[!htp]
\centering
	\begin{tikzpicture}
\begin{semilogyaxis} [grid = major,
    			xlabel = {epochs},
    			ylabel = {Training loss $L_1$},
    			width=0.45\textwidth,
    			height=0.4\textwidth,
    			/pgf/number format/1000 sep={},
    			ymax = 150			
		]
    		 \addplot[rwth1, very thick] table[x expr=\coordindex+1, y index=0] {graphs/mat12/Eichinger2020_02_DATA_UNIAXIAL-L1-2024-06-18_loss_history.txt};
\end{semilogyaxis}
\end{tikzpicture}
\begin{tikzpicture}
\begin{semilogyaxis} [grid = major,
    			xlabel = {epochs},
    			ylabel = {Training loss $L_2$},
    			width=0.45\textwidth,
    			height=0.4\textwidth,
    			/pgf/number format/1000 sep={},
    			ymax = 150    				
		]
    		 \addplot[rwth8, very thick] table[x expr=\coordindex+1, y index=0] {graphs/mat12/Eichinger2020_02_DATA_UNIAXIAL-L2-2024-06-18_loss_history.txt};
\end{semilogyaxis}
\end{tikzpicture}
\caption{Loss during training of the iCANN for the stripe specimen (uniaxial stress state). Left: Training with $L_1$ regularization, cf. Table~\ref{tab:weights_stripe}. Right: Training with $L_2$ regularization, cf. Table~\ref{tab:weights_stripe}. The loss is plotted on a logarithmic scale. In both cases, $4,000$ epochs were used.}
\label{fig:loss_stripe}
\end{figure}

\textbf{Discovery.} Our discovered weights are listed in Table~\ref{tab:weights_stripe}.
We investigated two types of regularization strategies: $L_1$ regularization, also known as Lasso regularization, and $L_2$ regularization, commonly referred to as ridge regression. The regularization parameters were selected based on prior experience with CANNs (see Section~\ref{sec:intro}).
For training our network, we exclusively used the experimental data for compressing the specimen (see Figure~\ref{fig:training_testing_stripe}).
For the uniaxial stripe specimen, we set the maximum number of epochs equal to $4,000$ epochs (see Figure~\ref{fig:loss_stripe}).
The experimental data for stretching remained unseen by the network and were utilized to assess its predictive capabilities. For both regularization strategies, we observed good agreement between the experimental data and the model's response for both training and testing phases.
\begin{table}[h]
\centering
\begin{tabular}{l | r r r}
 Weights 		& Regularized & $L_1=0.01$ & $L_2=0.001$ \\
\hline
 $w_{0,1}^\psi$ 			& No 	& 1.6990947 		& 1.2036339  \\
 $w_{0,2}^\psi$ 			& Yes 	& 0.10240719		& 0.07181329  \\
 $w_{1,1}^\psi$ 			& No 	& -3.5541244		& 1.2016658  \\
 $w_{1,2}^\psi$ 			& Yes 	& 0		& 0.3978735 \\
 $w_{\sigma,1}^\phi$		& Yes 	& 0		& 0 \\
 $w_{\sigma,2}^\phi$		& Yes 	& 0		& 0 \\ 
 $w_{\sigma,3}^\phi$		& No 	& 6.075556e-08		& 3.980602e-08  \\
 $w_{\sigma,4}^\phi$		& Yes 	& 0.02765466		& 0.03391496  \\
 $w_{\tau,1}^\phi$		& Yes 	& 0		& 0  \\
 $w_{\tau,2}^\phi$		& Yes 	& 0		& 0  \\ 
 $w_{\tau,3}^\phi$		& No 	& 3.5020828e-09		& 7.274134e-08  \\
 $w_{\tau,4}^\phi$		& Yes 	& 0		& 0.03408322  \\ 
 $\hat{w}_\eta$				& Yes 	& 0.43815053\tablefootnote{regularized with $L_2=0.001$}		& 0.26240048
\end{tabular}
\caption{Discovered weights for the stripe specimen (uniaxial stress state) with $L_1$ and $L_2$ regularization. The weights correspond to the feed-forward networks of the Helmholtz free energy~\eqref{eq:ffn_energy} and the homeostatic surface~\eqref{eq:ffn_potential}. The second column, `Regularized', indicates whether the regularization factor associated with the weight is zero (`No') or if regularization, $L_1$ or $L_2$, is applied (`Yes').}
\label{tab:weights_stripe}
\end{table}
We note that the experimental data exhibit a noticeable jump in the stress-time curve when the load is perturbed. In contrast, the jump in the network's prediction is several magnitudes smaller than that observed in the experiments. We attribute this discrepancy to two potential reasons:
(i) The perturbation in the experimental setup occurs in a quasi-instantaneous manner relative to the duration of the experiment and the growth and remodeling time. In the network architecture, we assume that the acceleration of displacement is negligibly small — a standard assumption. However, this may lead us to overlook inertia effects that could influence the stress response.
(ii) We lack detailed information about the testing device, which may not be designed to accurately measure such instantaneous perturbations, potentially resulting in overmodulation.

We assume an initially isotropic behavior, and thus, introduced an isochoric and a volumetric response.
As our network is interpretable, we can attribute the (elastic) shear modulus, $\mu$, and the (elastic) bulk modulus, $\kappa$, to the discovered weights at the (theoretical) initial state $\bar{\bm{C}}_e = \bm{C}_g = \bm{C} = \bm{I}$.
For the architecture chosen, we find $\mu = 2\,w_{1,2}^\psi\,\left(w_{1,1}^\psi\right)^2$ and $\kappa = 4\, w_{0,2}^\psi\,\left(w_{0,1}^\psi\right)^2$ (see \ref{app:moduli}).
It is interesting to note that the shear modulus turns out to be zero in case of Lasso regularization.
From a mechanical perspective, a non-negative shear modulus is acceptable; however, in reality, any material exhibits at least a small amount of shear resistance. Specifically, to perform quasi-static numerical simulations in an implicit manner, we encountered difficulties in inverting the stiffness matrix, as the corresponding material tangent is positive semi-definite.

This prompts us to consider the underlying reason for this behavior. Essentially, we are attempting to determine two independent material parameters using a one-dimensional stress-time curve. The solution to this problem is inherently non-unique. Since Lasso regularization penalizes all non-zero weights, the optimal loss can be achieved using only one material parameter, which may lead to the observed challenges. 
We discuss this issue in further detail in Section~\ref{sec:limits}.\newline

\textbf{Structural performance.} Lastly, we are interested in the performance of our discovered material model in a structural simulation.
Therefore, we implemented a material subroutine into the Finite Element software \textit{FEAP} \cite{feap_2020} and made the material subroutine accessible to the public.
For visualization, we utilized the open-source package \textit{ParaView} \cite{paraview}.
We discretized the problem using $4,580$ Q1 standard finite elements (see Figure~\ref{fig:mesh}), as we do not expect severe locking.
A similar boundary value problem for growth and remodeling was performed in \cite{holthusen2023}.
Thus, we kindly refer the reader to a detailed description of the boundary problem described therein.
As previously mentioned, the parameters obtained from Lasso regularization are not suitable for use in implicit structural simulations. 
Therefore, we use the parameters obtained using the $L_2$ regularization.
Figure~\ref{fig:structural_stripe} shows the stress contour plots of the second Piola-Kirchhoff stress, $\bm{S}$, at different time steps. 
Initially, as expected, the specimen contracts even though no external deformation is applied. From a qualitative perspective, our discovered model performs well during this phase.

However, over time, we observe a strain localization occurring within the edge elements. 
Ultimately, these two rows of elements become extensively stretched while the remaining elements in the center of the specimen contract, causing no convergence after $10.4$ [h]. 
This behavior is unexpected and does not align with experimental observations, indicating that certain limitations exist. 
These limitations will be discussed in further detail in Section~\ref{sec:limits}.
\begin{figure}[!htp]
\centering
	\begin{tikzpicture}
\begin{axis} [grid = major,
    			xlabel = {Time [h]},
    			ylabel = {Stress $S_{11}\ \left[\frac{\mu\text{N}}{\text{mm}^2}\right]$},
    			width=0.45\textwidth,
    			height=0.4\textwidth,
    			/pgf/number format/1000 sep={},
    			ymin = 0,
    			ymax = 15,    			
    			xmin = 0,
    			legend pos = south east    			
		]
\addplot[rwth1, very thick] table[x expr=\thisrowno{0},	y expr=\thisrowno{1}] {graphs/mat12/Eichinger2020_02_DATA_UNIAXIAL-L1-2024-06-18_TrainPredict.txt};
\addplot[rwth8, very thick, dashed] table[x expr=\thisrowno{0},	y expr=\thisrowno{1}] {graphs/mat12/Eichinger2020_02_DATA_UNIAXIAL-L2-2024-06-18_TrainPredict.txt};		
    		 \addplot[name path=mean, black, very thick] table[x expr=\thisrowno{0}, y expr=\thisrowno{1}] {graphs/Prepare_DATA/uniaxial_min.txt};
    		 \addplot[name path=upper, rwth6, thick] table[x expr=\thisrowno{0}, y expr=\thisrowno{3}] {graphs/Prepare_DATA/uniaxial_min.txt};
    		 \addplot[name path=lower, rwth6, thick] table[x expr=\thisrowno{0}, y expr=\thisrowno{4}] {graphs/Prepare_DATA/uniaxial_min.txt};
    \addplot [
        fill=rwth6, 
        fill opacity=0.3
    ] fill between [
        of=upper and lower
    ];  
\legend{Training $L_1$,Training $L_2$,Experiment}      		 
\end{axis}

\end{tikzpicture}
\begin{tikzpicture}
\begin{axis} [grid = major,
    			xlabel = {Time [h]},
    			ylabel = {Stress $S_{11}\ \left[\frac{\mu\text{N}}{\text{mm}^2}\right]$},
    			width=0.45\textwidth,
    			height=0.4\textwidth,
    			/pgf/number format/1000 sep={},
    			ymin = 0,
    			ymax = 15,
    			xmin = 0,
    			legend pos = south east
		]	
\addplot[rwth5, very thick, dashpattern1] table[x expr=\thisrowno{2},	y expr=\thisrowno{3}] {graphs/mat12/Eichinger2020_02_DATA_UNIAXIAL-L1-2024-06-18_TrainPredict.txt};	
\addplot[rwth9, very thick, dashpattern2] table[x expr=\thisrowno{2},	y expr=\thisrowno{3}] {graphs/mat12/Eichinger2020_02_DATA_UNIAXIAL-L2-2024-06-18_TrainPredict.txt};	
    		 \addplot[name path=mean, black, very thick] table[x expr=\thisrowno{0}, y expr=\thisrowno{1}] {graphs/Prepare_DATA/uniaxial_pos.txt};
    		 \addplot[name path=upper, rwth4, thick] table[x expr=\thisrowno{0}, y expr=\thisrowno{3}] {graphs/Prepare_DATA/uniaxial_pos.txt};
    		 \addplot[name path=lower, rwth4, thick] table[x expr=\thisrowno{0}, y expr=\thisrowno{4}] {graphs/Prepare_DATA/uniaxial_pos.txt};
    \addplot [
        fill=rwth4, 
        fill opacity=0.3
    ] fill between [
        of=upper and lower
    ];    
\legend{Testing $L_1$,Testing $L_2$,Experiment}	     		 
\end{axis}

\end{tikzpicture}
\caption{Discovered model for the stripe specimen (uniaxial stress state). The experimental data is taken form \cite{eichinger2020}. In both plots, the black curve represents the experimental mean from three specimens, while the shaded areas show the mean $\pm$ the standard error of the mean. The stress is plotted in terms of $\bm{S}$ in loading direction. Left: Training results for $L_1$ and $L_2$ regularization. In the experiment, the homeostatic force is reduced by $-10\%$ at $t=17$ [h]. Right: Testing results for $L_1$ and $L_2$ regularization, where the homeostatic force is increased by $+10\%$ at $t=17$ [h].}
\label{fig:training_testing_stripe}
\end{figure}
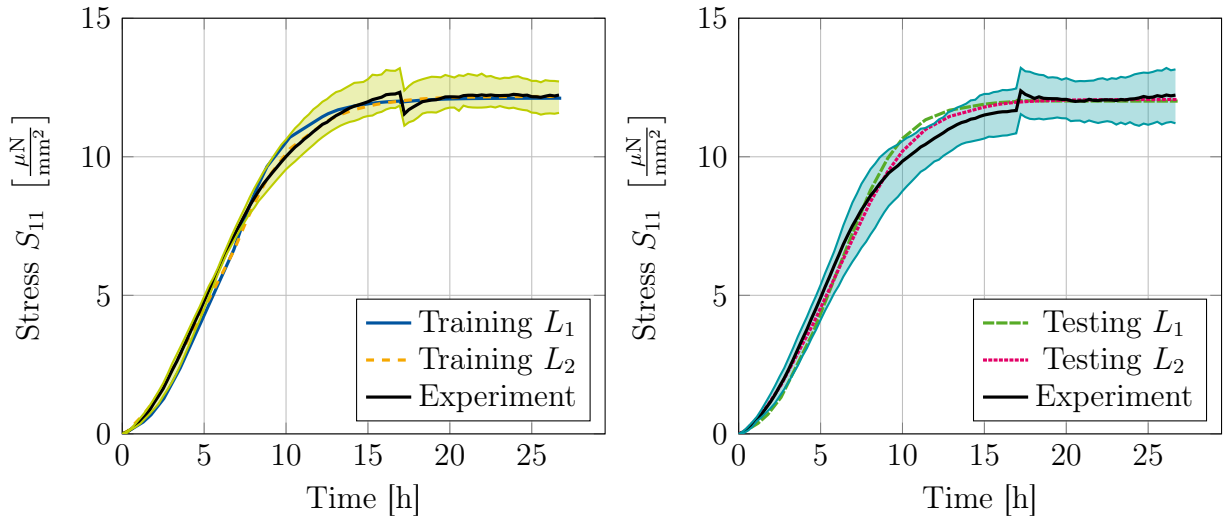
\begin{figure}[!htp]
	\centering
	\includegraphics[width=0.5\textwidth]{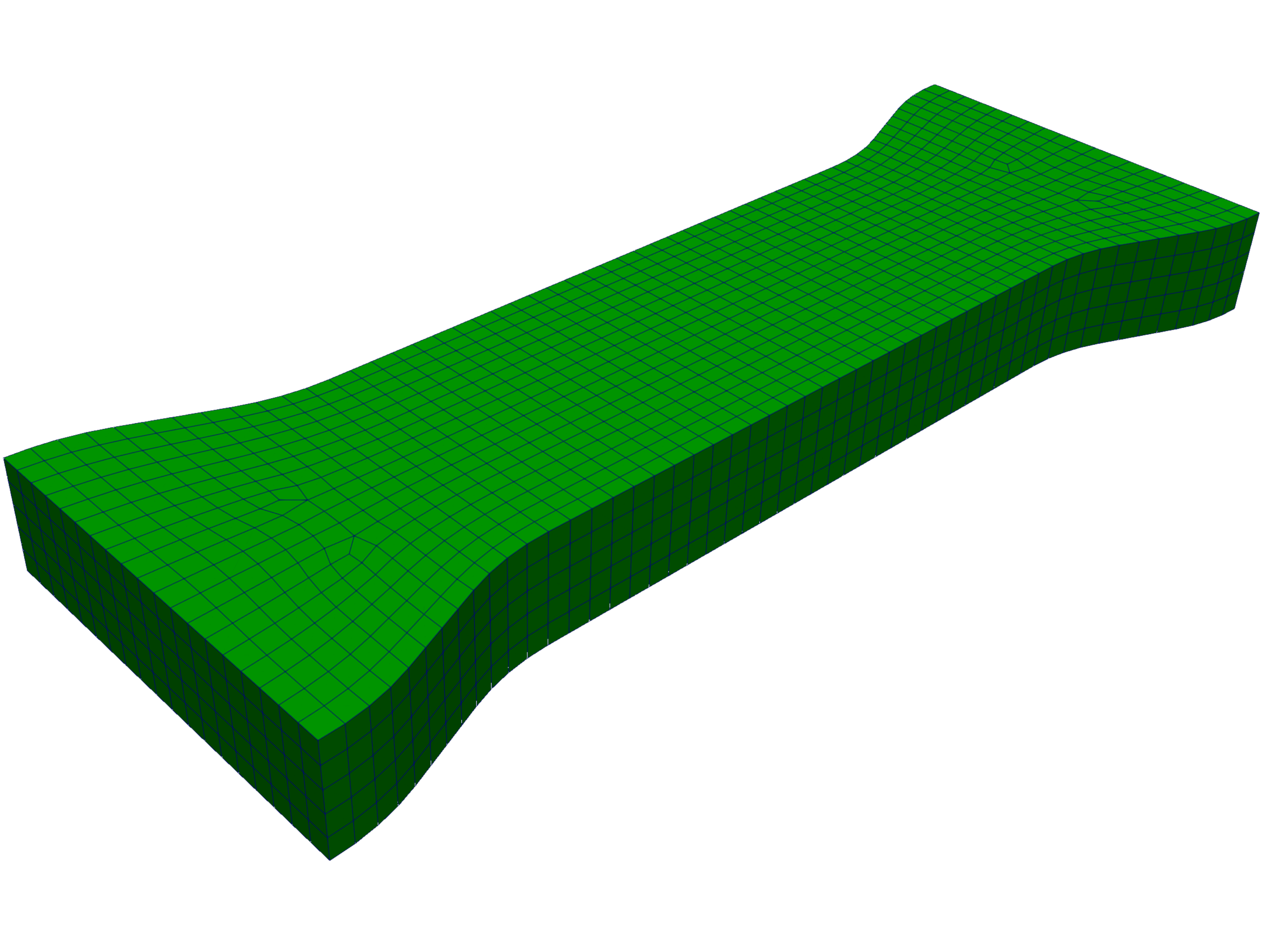}
	\caption{Finite element discretization with $4,580$ elements for the stripe specimen. The thickness of the specimen is equal to $4\ [\text{mm}]$, while the cross section area is equal to $40\ \left[\text{mm}^2\right]$ in the center of the specimen. For discretization in thickness direction, $5$ elements were used across the thickness.}
	\label{fig:mesh}
\end{figure}
\begin{figure}[!htp]
  \centering 

  \begin{subfigure}{.48\textwidth} 
    \centering 
    \begin{tikzpicture}
      \node[inner sep=0pt] (pic) at (0,0) {\includegraphics[width=\textwidth]{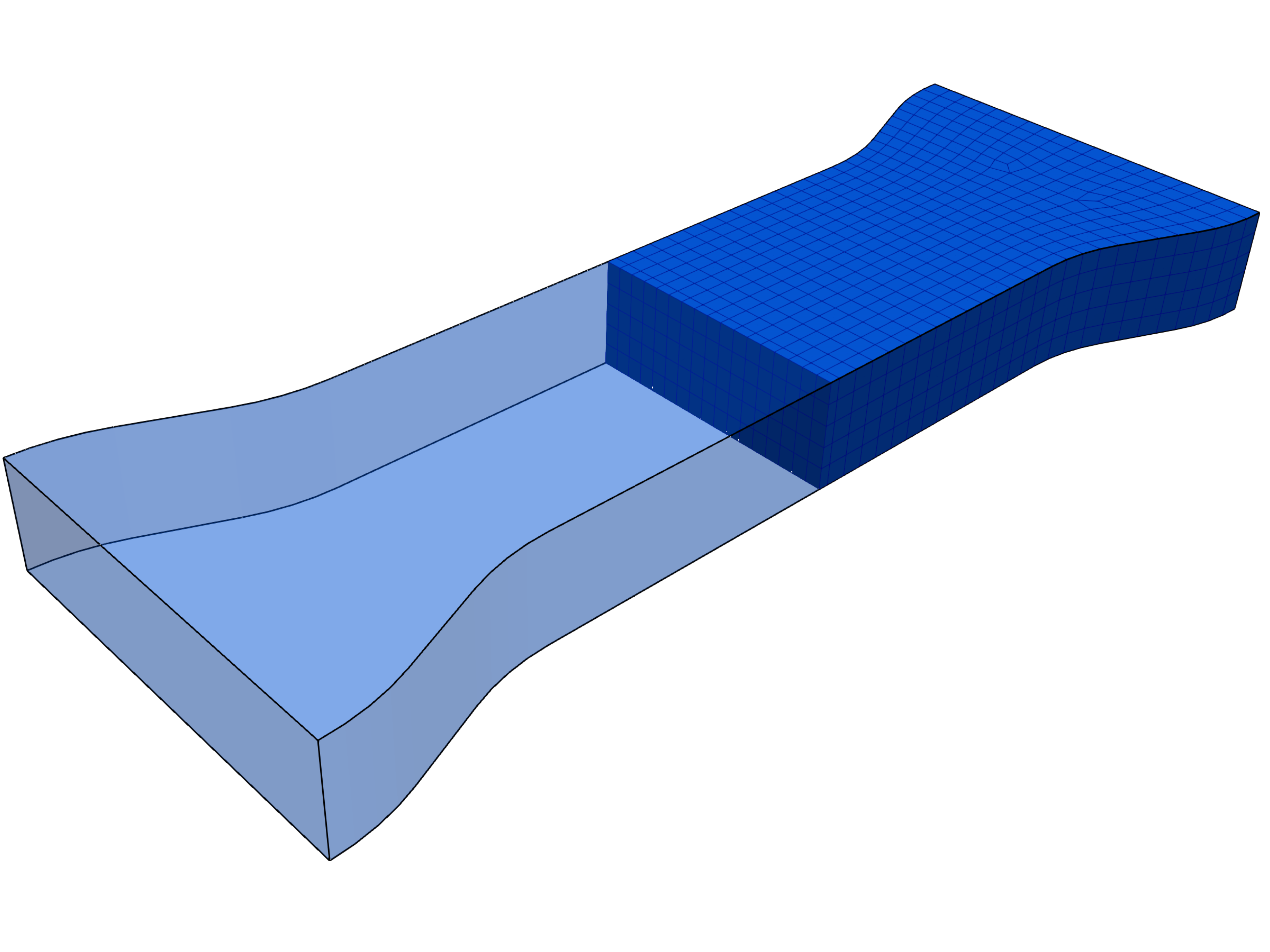}};
    \end{tikzpicture} 
    \caption{$t=0$ [h]}
  \end{subfigure}
  \begin{subfigure}{.48\textwidth} 
    \centering 
    \begin{tikzpicture}
      \node[inner sep=0pt] (pic) at (0,0) {\includegraphics[width=\textwidth]{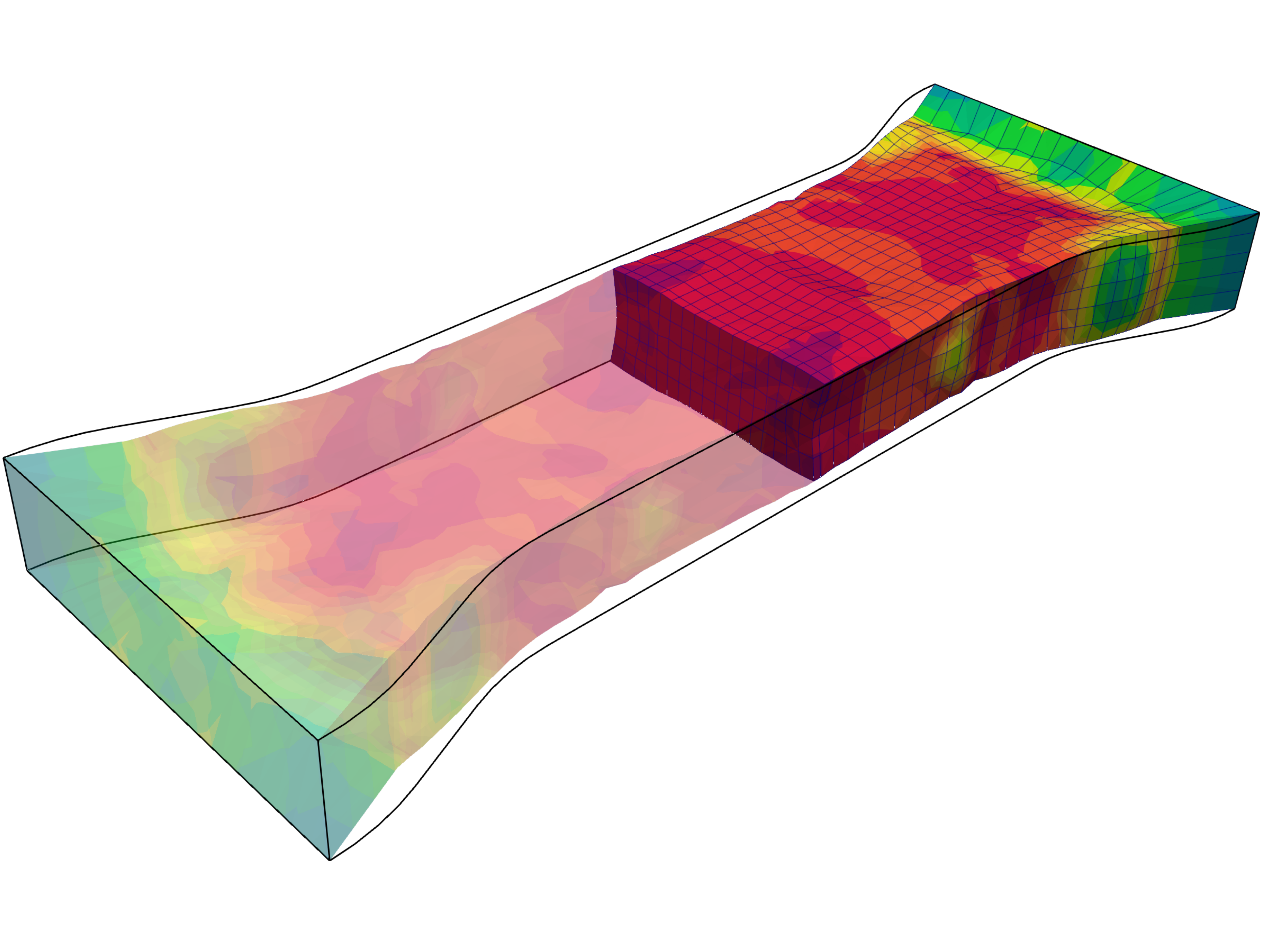}};
    \end{tikzpicture} 
    \caption{$t=6.5$ [h]}
  \end{subfigure}
  
  \begin{subfigure}{.48\textwidth} 
    \centering 
    \begin{tikzpicture}
      \node[inner sep=0pt] (pic) at (0,0) {\includegraphics[width=\textwidth]{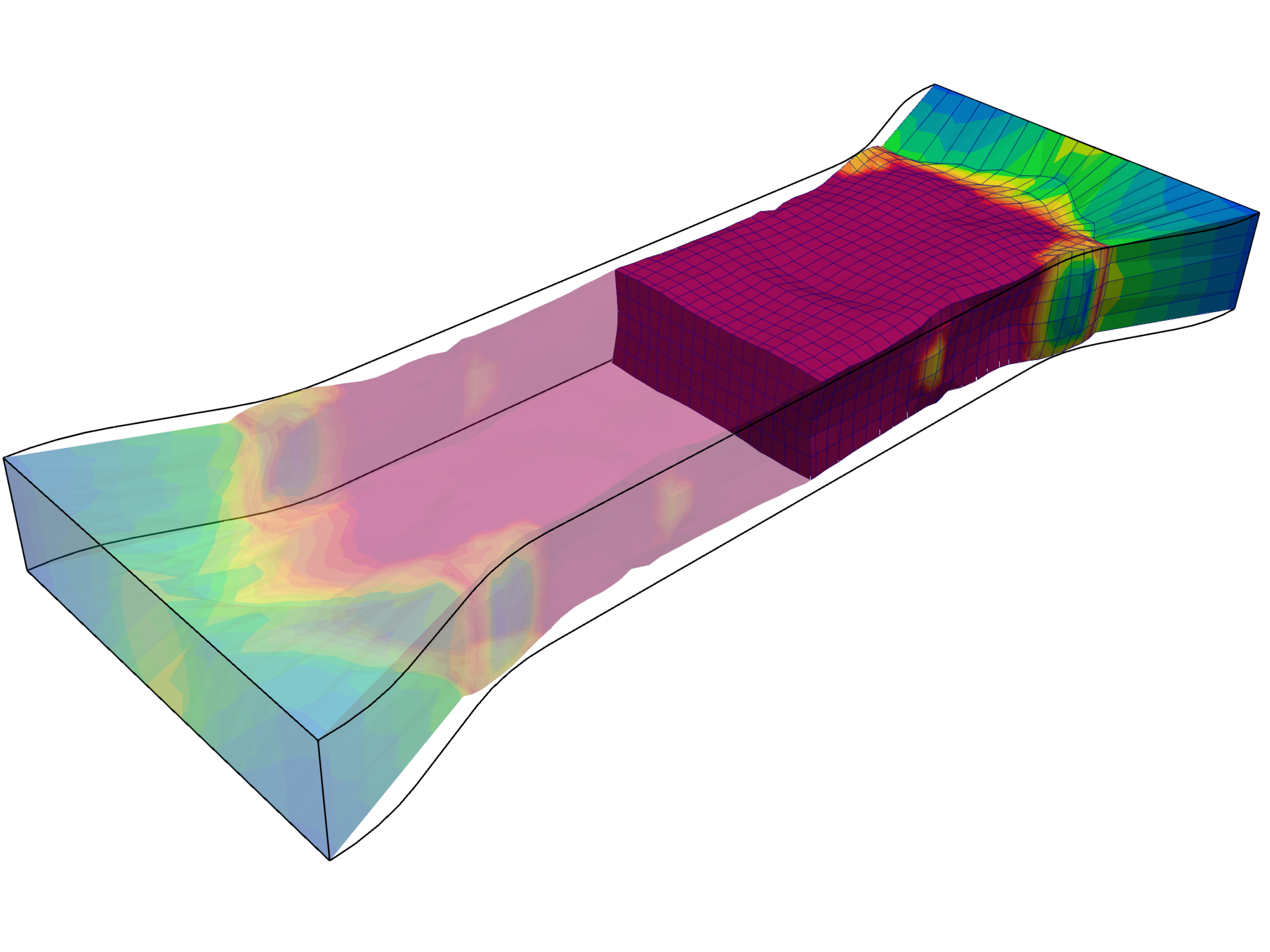}};
      \end{tikzpicture} 
    \caption{$t=8$ [h]}
  \end{subfigure}
  \begin{subfigure}{.48\textwidth} 
    \centering 
    \begin{tikzpicture}
      \node[inner sep=0pt] (pic) at (0,0) {\includegraphics[width=\textwidth]{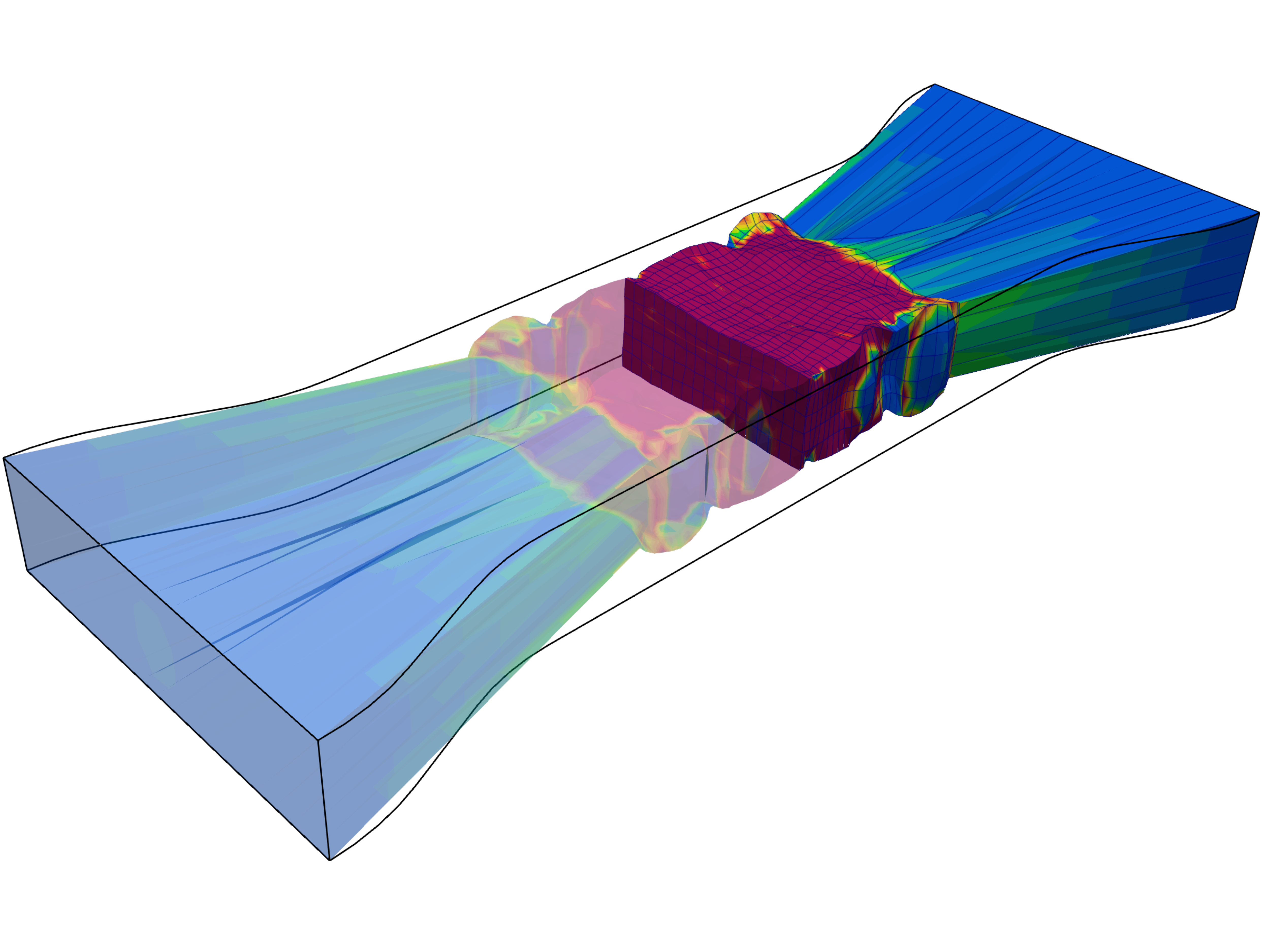}};
    \end{tikzpicture} 
    \caption{$t=10.4$ [h]}
  \end{subfigure}
  
  \vspace{2mm}
  
  \begin{subfigure}{\textwidth} 
    \centering 
    \begin{tikzpicture}
      \node[inner sep=0pt] (pic) at (0,0) {\includegraphics[height=5mm, width=40mm]{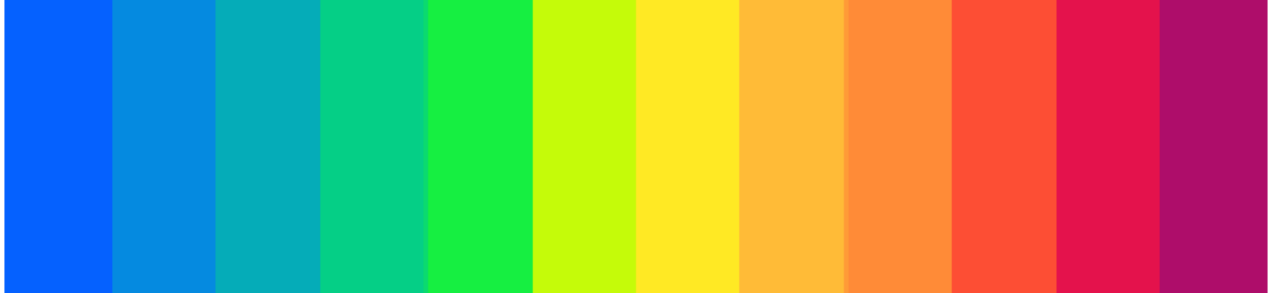}};
      \node[inner sep=0pt] (0)   at ($(pic.south)+(-2.22, 0.26)$)  {$0$};
      \node[inner sep=0pt] (1)   at ($(pic.south)+( 2.6, 0.26)$)  {$\geq 8.7$};
      \node[inner sep=0pt] (d)   at ($(pic.south)+(-3.80, 0.26)$)  {$S_{11}~\si{[\mu \N \per \mm\squared]}$};
      \node[inner sep=0pt] (d)   at ($(pic.south)+( 3.80, 0.26)$)  {\hphantom{$S_{11}~\si{[\mu \N \per \mm\squared]}$}};
    \end{tikzpicture} 
  \end{subfigure}
  
  \caption{Stress contour plots at different time steps for the discovered model of the stripe (see Figure~\ref{fig:training_testing_stripe}). The stress $S_{11}$ is plotted in longitudinal direction. On both edges, the specimen is fully clamped. The black lines present the undeformed shape of the specimen. No convergence was achieved after $t=10.4$ [h].} 
  \label{fig:structural_stripe}     
\end{figure}
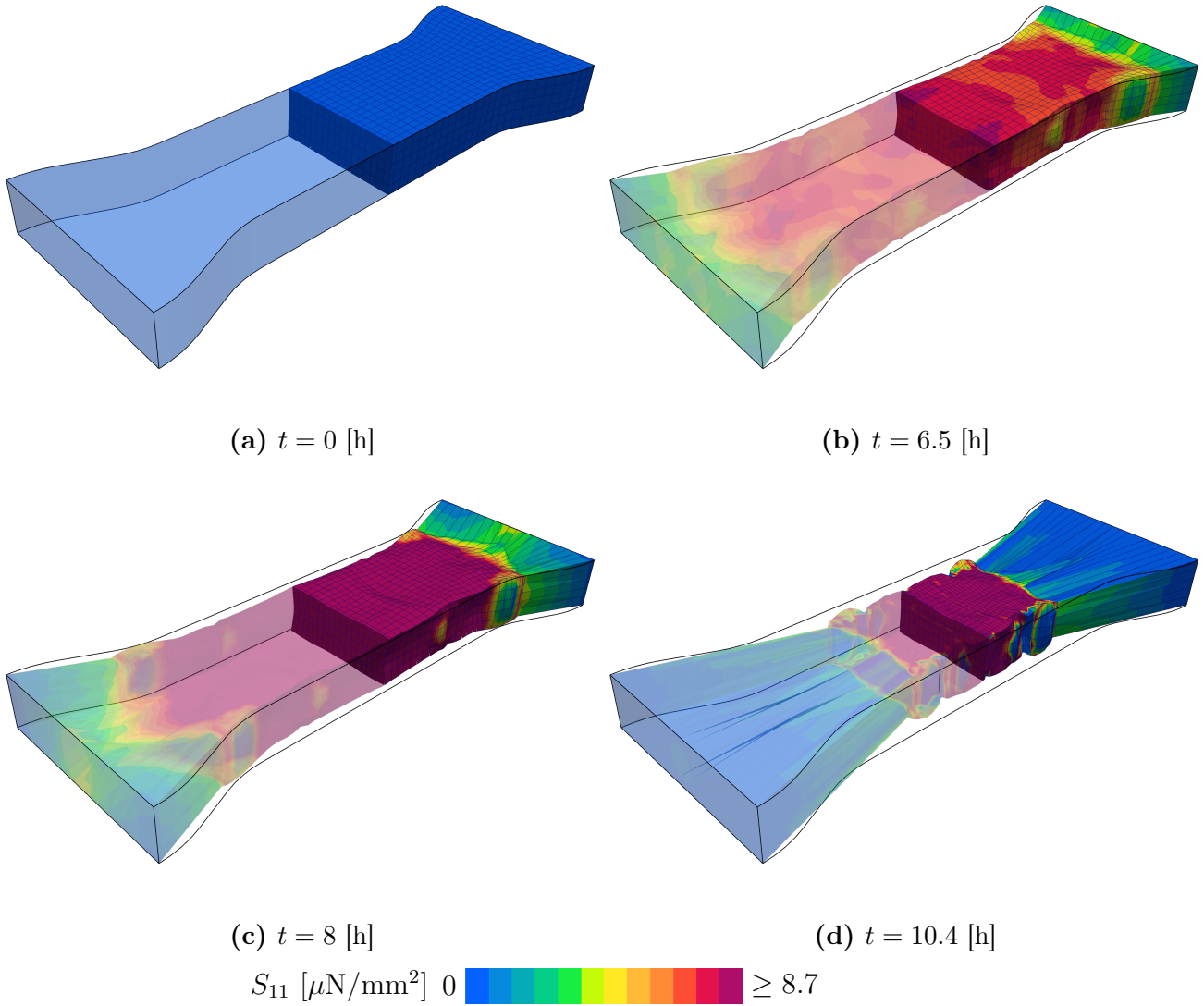
%
\subsection{Discovering a model for the cross specimen}
\label{sec:cross}
We proceed to evaluate the iCANN using the cross specimen, which is distinguished by two principal loading directions corresponding to the arms of the cross. These arms are clamped to facilitate displacement-driven loading. The experimental stress-time curves, represented in terms of the second Piola-Kirchhoff stress, are illustrated in Figures~\ref{fig:training_cross} through \ref{fig:testing_cross_semibiaxial_min} for both stretching and compressing the specimen.
Homeostasis is achieved after twenty-seven hours, during which no deformation is applied, signifying that $\bm{C}=\bm{I}$.
Two experimental setups are explored: (i) a biaxial loading scenario in which the forces in both principal loading directions are perturbed by $\pm 20\%$ at $t=27 [h]$, and (ii) a semi-biaxial loading scenario where the forces in one principal loading direction are perturbed by $t=27 [h]$, while the perpendicular loading direction remains undeformed. The resulting instantaneous deformations, expressed in terms of the right Cauchy-Green tensor, are summarized in Table~\ref{tab:loading_cross}.
\begin{table}[!htp]
\centering
\begin{tabular}{l | c c | c c}
  			& \multicolumn{2}{c}{Biaxial} & \multicolumn{2}{c}{Semi-biaxial} \\
 Loading		& $C_{11}\ [-]$ & $C_{22}\ [-]$ & $C_{11}\ [-]$ & $C_{22}\ [-]$ \\
\hline
 Stretching	 & 1.0044529e+00 & 1.0041029e+00 & 1.0046307e+00 & 1.0 \\
 Compressing & 9.9416188e-01 & 9.9379366e-01 & 9.9408145e-01 & 1.0 \\
\end{tabular}
\caption{Main diagonal components of the right Cauchy Green tensor, $\bm{C}$ at $t > 27\ [h]$. The loading scenarios correspond to the biaxial and semi-biaxial setups of the cross specimen. For both setups, the third loading direction remains always unchanged, i.e., $C_{33}=1.0\ [-]$, while we further assume zero stress in this very direction (see \ref{app:lagrange}).}
\label{tab:loading_cross}
\end{table}\newline
\textbf{Discovery.} The weights we identified are detailed in Table~\ref{tab:weights_cross}. In this investigation, we explored two regularization strategies: $L_1$ and $L_2$ regularization. For training our network, we utilized only the experimental data from the biaxial stretching of the specimen (see Figure~\ref{fig:training_cross}). The maximum number of training epochs was set to $8,000$ (see Figure~\ref{fig:loss_cross}).

The experimental data for the biaxial compression of the specimen, along with the data for both stretching and compressing in a semi-biaxial manner, were not presented to the network during training; instead, they were reserved for evaluating the model's predictive capabilities. For both regularization approaches, we observe a good agreement between the experimental results and the model's responses during both the training and testing phases.

Despite this, the model's predictions for the test data under semi-biaxial compression, as illustrated in Figure~\ref{fig:testing_cross_semibiaxial_min}, exhibit notable discrepancies. Several factors may contribute to this deviation.
Firstly, the continuum mechanical formulation inherently predicts the same homeostatic stress in this homogeneous loading scenario across both loading directions. This is a fundamental assumption in the isotropic formulation of the homeostatic surface. Incorporating structural tensors could allow for direction-specific homeostatic stresses, potentially improving the model's accuracy.
Whether this is really necessary is out of the scope of this contribution.
Secondly, the standard error of the mean between the two loading directions in this particular setup is relatively high. The extent of homeostatic stresses in each direction may also be influenced by the inherent uncertainty in the experimental data — a factor that is not accounted for in the deterministic approach employed here.
Lastly, as already discussed in \cite{eichinger2020}, the applied deformation may trigger proliferation, resulting in a shift in the homeostatic stress that is not considered in the current model.

Notably, a significant jump is evident in the experimental data, which is not mirrored in the model response. We attribute this discrepancy to the same factors discussed previously concerning the stripe specimen.

We remain focused on identifying the (elastic) material parameters. However, in this case, both regularization methods resulted in a shear modulus of zero, rendering implicit structural simulation impractical. This outcome is somewhat counterintuitive; one might expect that more complex loading scenarios (such as biaxial loading compared to uniaxial) would provide additional information, leading to a more accurate model discovery.

One possible explanation is that when an isotropic material is stretched or compressed biaxially with similar magnitudes in the perpendicular loading directions (as shown in Table~\ref{tab:loading_cross}), the stress response remains consistent. Consequently, this setup does not impart additional information regarding the material's underlying behavior.

Thus, one might consider using either the stretching or compressing semi-biaxial data for training. However, a closer examination of the experimental data in Figures~\ref{fig:testing_cross_semibiaxial_pos} through \ref{fig:testing_cross_semibiaxial_min} reveals no clear distinction between the two stress components, $S_{11}$ and $S_{22}$, even after the perturbation is applied. This lack of differentiation could be attributed to the perturbation being insufficient to elicit a clear response given the non-proportional loading conditions.

We hypothesize that extended non-proportional loading may better reveal the nonlinear material behavior and potentially discover both material parameters in question. This issue concerning data sparsity will be discussed in more detail in Section~\ref{sec:limits}.
\begin{table}[!htp]
\centering
\begin{tabular}{l | r r r}
 Weights 		& Regularized & $L_1=0.01$ & $L_2=0.001$ \\
\hline
 $w_{0,1}^\psi$ 			& No 	& 2.168842  			&  3.1134224 \\
 $w_{0,2}^\psi$ 			& Yes 	& 0.27726683			&  0.36447218 \\
 $w_{1,1}^\psi$ 			& No 	& 1.3061364			& -0.2970376  \\
 $w_{1,2}^\psi$ 			& Yes 	& 0					&  0 \\
 $w_{\sigma,1}^\phi$		& Yes 	& 0					&  0 \\
 $w_{\sigma,2}^\phi$		& Yes 	& 0					&  0 \\ 
 $w_{\sigma,3}^\phi$		& No 	& 6.806422e-08		& 0  \\
 $w_{\sigma,4}^\phi$		& Yes 	& 0.01459178			& 0.01457657  \\
 $w_{\tau,1}^\phi$		& Yes 	& 0					& 0  \\
 $w_{\tau,2}^\phi$		& Yes 	& 0					& 0  \\ 
 $w_{\tau,3}^\phi$		& No 	& 2.7375126e-08		& 9.1654684e-08  \\
 $w_{\tau,4}^\phi$		& Yes 	& 0.0012281			& 0.01357422  \\ 
 $\hat{w}_\eta$				& Yes 	& 0.27317414\tablefootnote{regularized with $L_2=0.001$}		& 0.49805772
\end{tabular}
\caption{Discovered weights for the cross specimen (biaxial stress state) with $L_1$ and $L_2$ regularization. The weights correspond to the feed-forward networks of the Helmholtz free energy~\eqref{eq:ffn_energy} and the homeostatic surface~\eqref{eq:ffn_potential}. The second column, `Regularized', indicates whether the regularization factor associated with the weight is zero (`No') or if regularization, $L_1$ or $L_2$, is applied (`Yes').}
\label{tab:weights_cross}
\end{table}
\begin{figure}[!htp]
\centering
	\begin{tikzpicture}
\begin{semilogyaxis} [grid = major,
    			xlabel = {epochs},
    			ylabel = {Training loss $L_1$},
    			width=0.45\textwidth,
    			height=0.4\textwidth,
    			/pgf/number format/1000 sep={},
    			ymax = 1050			
		]
    		 \addplot[rwth1, very thick] table[x expr=\coordindex+1, y index=0] {graphs/mat12/Eichinger2020_02_DATA-L1-2024-06-18_loss_history.txt};
\end{semilogyaxis}
\end{tikzpicture}
\begin{tikzpicture}
\begin{semilogyaxis} [grid = major,
    			xlabel = {epochs},
    			ylabel = {Training loss $L_2$},
    			width=0.45\textwidth,
    			height=0.4\textwidth,
    			/pgf/number format/1000 sep={},
    			ymax = 1050    				
		]
    		 \addplot[rwth8, very thick] table[x expr=\coordindex+1, y index=0] {graphs/mat12/Eichinger2020_02_DATA-L2-2024-06-18_loss_history.txt};
\end{semilogyaxis}
\end{tikzpicture}
\caption{Loss during training of the iCANN for the cross specimen (biaxial stress state). Left: Training with $L_1$ regularization, cf. Table~\ref{tab:weights_cross}. Right: Training with $L_2$ regularization, cf. Table~\ref{tab:weights_cross}. The loss is plotted on a logarithmic scale. In both cases, $8,000$ epochs were used.}
\label{fig:loss_cross}
\end{figure}
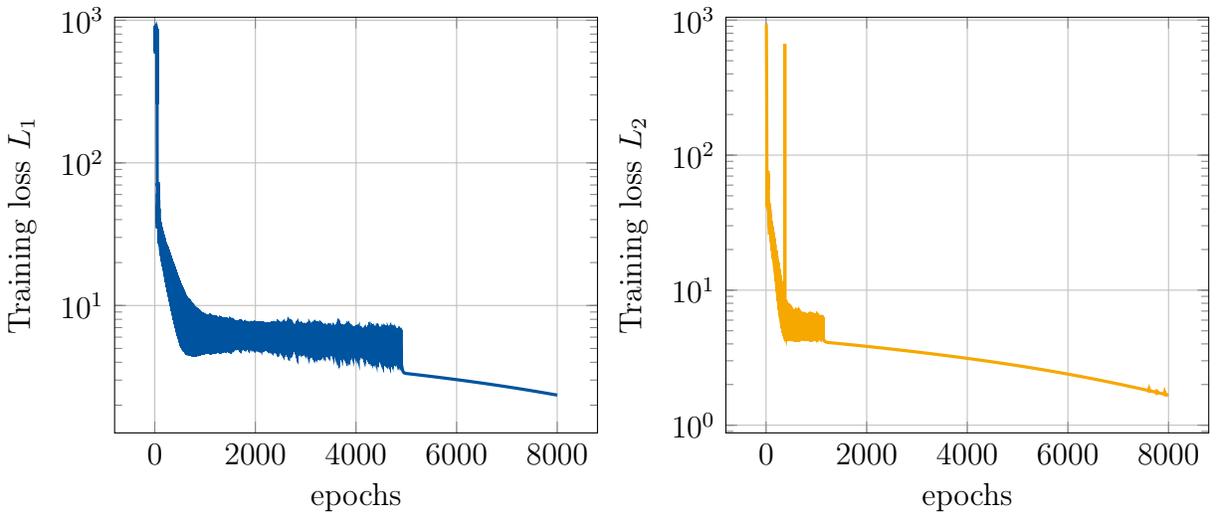
\begin{figure}[!htp]
\centering
	\begin{tikzpicture}
\begin{axis} [grid = major,
    			xlabel = {Time [h]},
    			ylabel = {Stress $S_{11}\ \left[\frac{\mu\text{N}}{\text{mm}^2}\right]$},
    			width=0.45\textwidth,
    			height=0.4\textwidth,
    			/pgf/number format/1000 sep={},
    			ymin = 0,
    			ymax = 28,
    			xmin = 0,
    			legend pos=south east
		]
    \addplot[rwth1, very thick] table[x expr=\thisrowno{0},	y expr=\thisrowno{1}] {graphs/mat12/Eichinger2020_02_DATA-L1-2024-06-18_TrainPredict.txt};  	
    \addplot[rwth8, very thick, dashed] table[x expr=\thisrowno{0},	y expr=\thisrowno{1}] {graphs/mat12/Eichinger2020_02_DATA-L2-2024-06-18_TrainPredict.txt};
    		 \addplot[name path=mean, black, very thick] table[x expr=\thisrowno{0}, y expr=\thisrowno{1}] {graphs/Prepare_DATA/biaxial_pos.txt};
    		 \addplot[name path=upper, rwth4, thick] table[x expr=\thisrowno{0}, y expr=\thisrowno{3}] {graphs/Prepare_DATA/biaxial_pos.txt};
    		 \addplot[name path=lower, rwth4, thick] table[x expr=\thisrowno{0}, y expr=\thisrowno{4}] {graphs/Prepare_DATA/biaxial_pos.txt};
    \addplot [
        fill=rwth4, 
        fill opacity=0.3
    ] fill between [
        of=upper and lower
    ];
\legend{Training $L_1$,Training $L_2$,Experiment}
\end{axis}

\end{tikzpicture}
\begin{tikzpicture}
\begin{axis} [grid = major,
    			xlabel = {Time [h]},
    			ylabel = {Stress $S_{22}\ \left[\frac{\mu\text{N}}{\text{mm}^2}\right]$},
    			width=0.45\textwidth,
    			height=0.4\textwidth,
    			/pgf/number format/1000 sep={},
    			ymin = 0,
    			ymax = 28,    			
    			xmin = 0,
    			legend pos=south east
		]
    \addplot[rwth1, very thick] table[x expr=\thisrowno{0},	y expr=\thisrowno{2}] {graphs/mat12/Eichinger2020_02_DATA-L1-2024-06-18_TrainPredict.txt};	
    \addplot[rwth8, very thick, dashed] table[x expr=\thisrowno{0},	y expr=\thisrowno{2}] {graphs/mat12/Eichinger2020_02_DATA-L2-2024-06-18_TrainPredict.txt};  	
    		 \addplot[name path=mean, black, very thick] table[x expr=\thisrowno{0}, y expr=\thisrowno{2}] {graphs/Prepare_DATA/biaxial_pos.txt};
    		 \addplot[name path=upper, rwth4, thick] table[x expr=\thisrowno{0}, y expr=\thisrowno{5}] {graphs/Prepare_DATA/biaxial_pos.txt};
    		 \addplot[name path=lower, rwth4, thick] table[x expr=\thisrowno{0}, y expr=\thisrowno{6}] {graphs/Prepare_DATA/biaxial_pos.txt};
    \addplot [
        fill=rwth4, 
        fill opacity=0.3
    ] fill between [
        of=upper and lower
    ];
\legend{Training $L_1$,Training $L_2$,Experiment}	 
\end{axis}

\end{tikzpicture}
\caption{\textbf{Training.} Discovered model for the cross specimen (biaxial stress state) for $L_1$ and $L_2$ regularization. The homeostatic force is increased by $+10\%$ at $t=27$ [h]. The experimental data is taken form \cite{eichinger2020}. In both plots, the black curve represents the experimental mean from three specimens, while the shaded areas show the mean $\pm$ the standard error of the mean. The stress is plotted in terms of $\bm{S}$ in both loading directions. Left: Stress in the first loading direction. Right: Stress in the perpendicular loading direction.}
\label{fig:training_cross}
\end{figure}
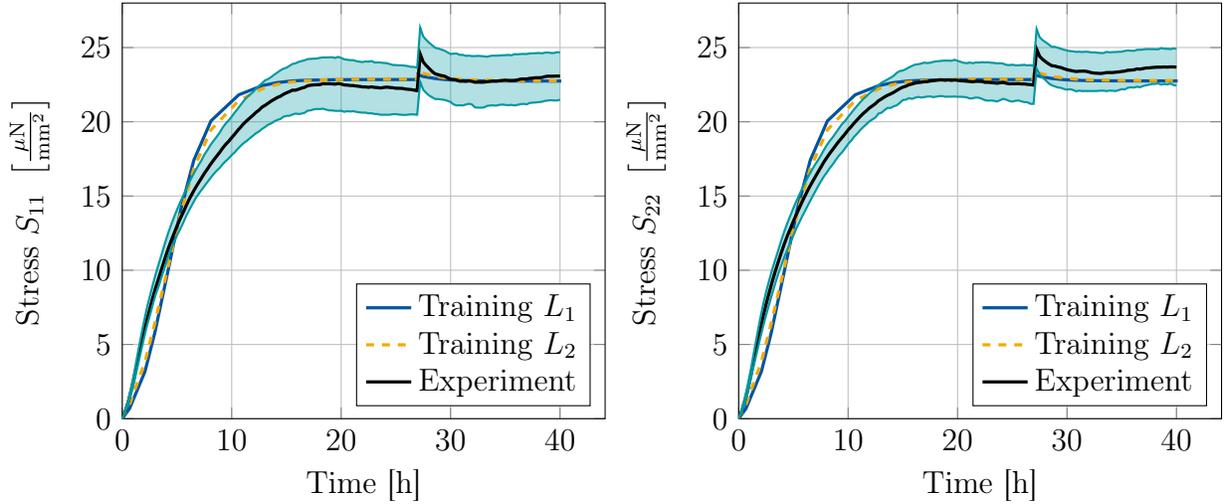
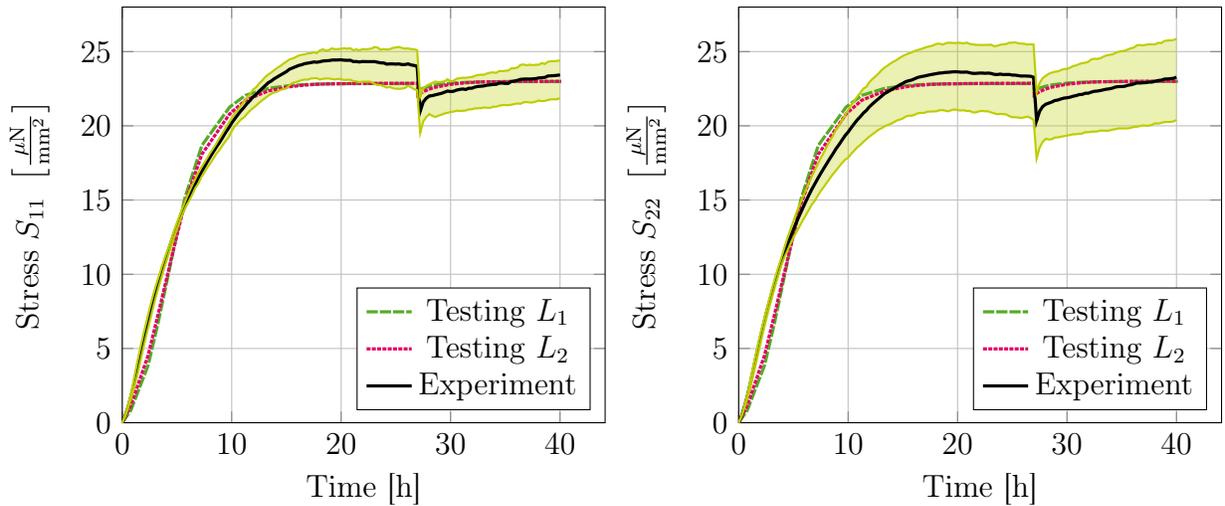
\begin{figure}[!htp]
\centering
	\begin{tikzpicture}
\begin{axis} [grid = major,
    			xlabel = {Time [h]},
    			ylabel = {Stress $S_{11}\ \left[\frac{\mu\text{N}}{\text{mm}^2}\right]$},
    			width=0.45\textwidth,
    			height=0.4\textwidth,
    			/pgf/number format/1000 sep={},
    			ymin = 0,
    			ymax = 28,    			
    			xmin = 0,
    			legend pos=south east
		]
    \addplot[rwth5, very thick, dashpattern1] table[x expr=\thisrowno{3},	y expr=\thisrowno{4}] {graphs/mat12/Eichinger2020_02_DATA-L1-2024-06-18_TrainPredict.txt};	
    \addplot[rwth9, very thick, dashpattern2] table[x expr=\thisrowno{3},	y expr=\thisrowno{4}] {graphs/mat12/Eichinger2020_02_DATA-L2-2024-06-18_TrainPredict.txt}; 	
    		 \addplot[name path=mean, black, very thick] table[x expr=\thisrowno{0}, y expr=\thisrowno{1}] {graphs/Prepare_DATA/biaxial_min.txt};
    		 \addplot[name path=upper, rwth6, thick] table[x expr=\thisrowno{0}, y expr=\thisrowno{3}] {graphs/Prepare_DATA/biaxial_min.txt};
    		 \addplot[name path=lower, rwth6, thick] table[x expr=\thisrowno{0}, y expr=\thisrowno{4}] {graphs/Prepare_DATA/biaxial_min.txt};
    \addplot [
        fill=rwth6, 
        fill opacity=0.3
    ] fill between [
        of=upper and lower
    ];    
\legend{Testing $L_1$,Testing $L_2$,Experiment}    		 
\end{axis}

\end{tikzpicture}
\begin{tikzpicture}
\begin{axis} [grid = major,
    			xlabel = {Time [h]},
    			ylabel = {Stress $S_{22}\ \left[\frac{\mu\text{N}}{\text{mm}^2}\right]$},
    			width=0.45\textwidth,
    			height=0.4\textwidth,
    			/pgf/number format/1000 sep={},
    			ymin = 0,
    			ymax = 28,    			
    			xmin = 0,
    			legend pos=south east
		]
    \addplot[rwth5, very thick, dashpattern1] table[x expr=\thisrowno{3},	y expr=\thisrowno{5}] {graphs/mat12/Eichinger2020_02_DATA-L1-2024-06-18_TrainPredict.txt};	
    \addplot[rwth9, very thick, dashpattern2] table[x expr=\thisrowno{3},	y expr=\thisrowno{5}] {graphs/mat12/Eichinger2020_02_DATA-L2-2024-06-18_TrainPredict.txt};	
    		 \addplot[name path=mean, black, very thick] table[x expr=\thisrowno{0}, y expr=\thisrowno{2}] {graphs/Prepare_DATA/biaxial_min.txt};
    		 \addplot[name path=upper, rwth6, thick] table[x expr=\thisrowno{0}, y expr=\thisrowno{5}] {graphs/Prepare_DATA/biaxial_min.txt};
    		 \addplot[name path=lower, rwth6, thick] table[x expr=\thisrowno{0}, y expr=\thisrowno{6}] {graphs/Prepare_DATA/biaxial_min.txt};
    \addplot [
        fill=rwth6, 
        fill opacity=0.3
    ] fill between [
        of=upper and lower
    ];   
\legend{Testing $L_1$,Testing $L_2$,Experiment}     		 
\end{axis}

\end{tikzpicture}
\caption{\textbf{Testing.} Discovered model prediction for the cross specimen (biaxial stress state) for $L_1$ and $L_2$ regularization. The homeostatic force is decreased by $-10\%$ at $t=27$ [h]. The experimental data is taken form \cite{eichinger2020}. In both plots, the black curve represents the experimental mean from three specimens, while the shaded areas show the mean $\pm$ the standard error of the mean. The stress is plotted in terms of $\bm{S}$ in both loading directions. Left: Stress in the first loading direction. Right: Stress in the perpendicular loading direction.}
\label{fig:testing_cross_biaxial}
\end{figure}
\begin{figure}[!htp]
\centering
	\begin{tikzpicture}
\begin{axis} [grid = major,
    			xlabel = {Time [h]},
    			ylabel = {Stress $S_{11}\ \left[\frac{\mu\text{N}}{\text{mm}^2}\right]$},
    			width=0.45\textwidth,
    			height=0.4\textwidth,
    			/pgf/number format/1000 sep={},
    			ymin = 0,
    			ymax = 29,
    			xmin = 0,
    			legend pos=south east    			
		]
    \addplot[rwth5, very thick, dashpattern1] table[x expr=\thisrowno{6},	y expr=\thisrowno{7}] {graphs/mat12/Eichinger2020_02_DATA-L1-2024-06-18_TrainPredict.txt};	
    \addplot[rwth9, very thick, dashpattern2] table[x expr=\thisrowno{6},	y expr=\thisrowno{7}] {graphs/mat12/Eichinger2020_02_DATA-L2-2024-06-18_TrainPredict.txt};	
    		 \addplot[name path=mean, black, very thick] table[x expr=\thisrowno{0}, y expr=\thisrowno{1}] {graphs/Prepare_DATA/semibiaxial_pos.txt};
    		 \addplot[name path=upper, rwth4, thick] table[x expr=\thisrowno{0}, y expr=\thisrowno{3}] {graphs/Prepare_DATA/semibiaxial_pos.txt};
    		 \addplot[name path=lower, rwth4, thick] table[x expr=\thisrowno{0}, y expr=\thisrowno{4}] {graphs/Prepare_DATA/semibiaxial_pos.txt};
    \addplot [
        fill=rwth4, 
        fill opacity=0.3
    ] fill between [
        of=upper and lower
    ];    
\legend{Testing $L_1$,Testing $L_2$,Experiment}    		 
\end{axis}

\end{tikzpicture}
\begin{tikzpicture}
\begin{axis} [grid = major,
    			xlabel = {Time [h]},
    			ylabel = {Stress $S_{22}\ \left[\frac{\mu\text{N}}{\text{mm}^2}\right]$},
    			width=0.45\textwidth,
    			height=0.4\textwidth,
    			/pgf/number format/1000 sep={},
    			ymin = 0,
    			ymax = 29,    			
    			xmin = 0,
    			legend pos=south east
		]
    \addplot[rwth5, very thick, dashpattern1] table[x expr=\thisrowno{6},	y expr=\thisrowno{8}] {graphs/mat12/Eichinger2020_02_DATA-L1-2024-06-18_TrainPredict.txt};	
    \addplot[rwth9, very thick, dashpattern2] table[x expr=\thisrowno{6},	y expr=\thisrowno{8}] {graphs/mat12/Eichinger2020_02_DATA-L2-2024-06-18_TrainPredict.txt};	
    		 \addplot[name path=mean, black, very thick] table[x expr=\thisrowno{0}, y expr=\thisrowno{2}] {graphs/Prepare_DATA/semibiaxial_pos.txt};
    		 \addplot[name path=upper, rwth4, thick] table[x expr=\thisrowno{0}, y expr=\thisrowno{5}] {graphs/Prepare_DATA/semibiaxial_pos.txt};
    		 \addplot[name path=lower, rwth4, thick] table[x expr=\thisrowno{0}, y expr=\thisrowno{6}] {graphs/Prepare_DATA/semibiaxial_pos.txt};
    \addplot [
        fill=rwth4, 
        fill opacity=0.3
    ] fill between [
        of=upper and lower
    ];    
\legend{Testing $L_1$,Testing $L_2$,Experiment}    		 
\end{axis}

\end{tikzpicture}
\caption{\textbf{Testing.} Discovered model prediction for the cross specimen (semi-biaxial stress state) for $L_1$ and $L_2$ regularization. The experimental data is taken form \cite{eichinger2020}. In both plots, the black curve represents the experimental mean from three specimens, while the shaded areas show the mean $\pm$ the standard error of the mean. The stress is plotted in terms of $\bm{S}$ in both loading directions. Left: Stress in the principal loading direction, where the homeostatic force is increased by $+10\%$ at $t=27$ [h]. Right: Stress in the perpendicular loading direction, which is not actively perturbed by the experimental device, though stress may still be affected indirectly.}
\label{fig:testing_cross_semibiaxial_pos}
\end{figure}
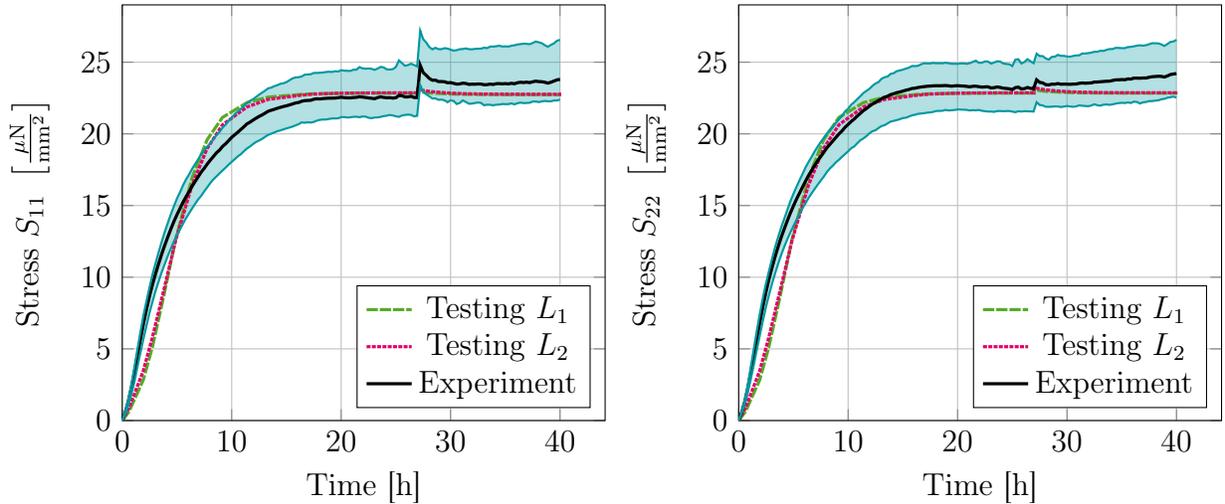
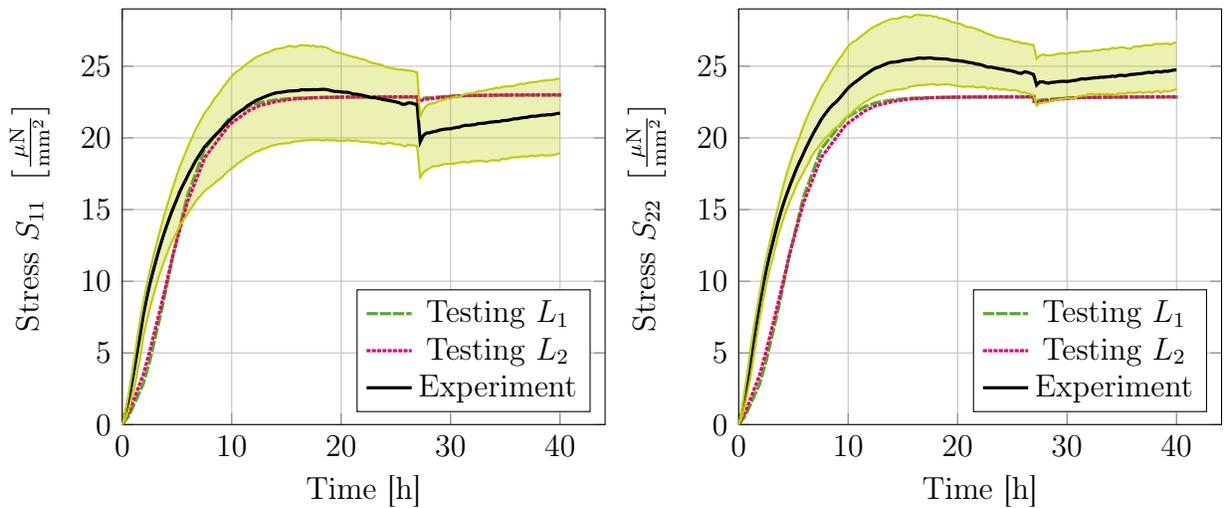
\begin{figure}[!htp]
\centering
	\begin{tikzpicture}
\begin{axis} [grid = major,
    			xlabel = {Time [h]},
    			ylabel = {Stress $S_{11}\ \left[\frac{\mu\text{N}}{\text{mm}^2}\right]$},
    			width=0.45\textwidth,
    			height=0.4\textwidth,
    			/pgf/number format/1000 sep={},
    			ymin = 0,
    			ymax = 29,    			
    			xmin = 0,
    			legend pos=south east
		]
    \addplot[rwth5, very thick, dashpattern1] table[x expr=\thisrowno{9},	y expr=\thisrowno{10}] {graphs/mat12/Eichinger2020_02_DATA-L1-2024-06-18_TrainPredict.txt};	
    \addplot[rwth9, very thick, dashpattern2] table[x expr=\thisrowno{9},	y expr=\thisrowno{10}] {graphs/mat12/Eichinger2020_02_DATA-L2-2024-06-18_TrainPredict.txt};	
    		 \addplot[name path=mean, black, very thick] table[x expr=\thisrowno{0}, y expr=\thisrowno{1}] {graphs/Prepare_DATA/semibiaxial_min.txt};
    		 \addplot[name path=upper, rwth6, thick] table[x expr=\thisrowno{0}, y expr=\thisrowno{3}] {graphs/Prepare_DATA/semibiaxial_min.txt};
    		 \addplot[name path=lower, rwth6, thick] table[x expr=\thisrowno{0}, y expr=\thisrowno{4}] {graphs/Prepare_DATA/semibiaxial_min.txt};
    \addplot [
        fill=rwth6, 
        fill opacity=0.3
    ] fill between [
        of=upper and lower
    ];   
\legend{Testing $L_1$,Testing $L_2$,Experiment}     		 
\end{axis}

\end{tikzpicture}
\begin{tikzpicture}
\begin{axis} [grid = major,
    			xlabel = {Time [h]},
    			ylabel = {Stress $S_{22}\ \left[\frac{\mu\text{N}}{\text{mm}^2}\right]$},
    			width=0.45\textwidth,
    			height=0.4\textwidth,
    			/pgf/number format/1000 sep={},
    			ymin = 0,
    			ymax = 29,    			
    			xmin = 0,
    			legend pos=south east
		]
    \addplot[rwth5, very thick, dashpattern1] table[x expr=\thisrowno{9},	y expr=\thisrowno{11}] {graphs/mat12/Eichinger2020_02_DATA-L1-2024-06-18_TrainPredict.txt};	
    \addplot[rwth9, very thick, dashpattern2] table[x expr=\thisrowno{9},	y expr=\thisrowno{11}] {graphs/mat12/Eichinger2020_02_DATA-L2-2024-06-18_TrainPredict.txt};	
    		 \addplot[name path=mean, black, very thick] table[x expr=\thisrowno{0}, y expr=\thisrowno{2}] {graphs/Prepare_DATA/semibiaxial_min.txt};
    		 \addplot[name path=upper, rwth6, thick] table[x expr=\thisrowno{0}, y expr=\thisrowno{5}] {graphs/Prepare_DATA/semibiaxial_min.txt};
    		 \addplot[name path=lower, rwth6, thick] table[x expr=\thisrowno{0}, y expr=\thisrowno{6}] {graphs/Prepare_DATA/semibiaxial_min.txt};
    \addplot [
        fill=rwth6, 
        fill opacity=0.3
    ] fill between [
        of=upper and lower
    ];    	
\legend{Testing $L_1$,Testing $L_2$,Experiment}    	 
\end{axis}

\end{tikzpicture}
\caption{\textbf{Testing.} Discovered model prediction for the cross specimen (semi-biaxial stress state) for $L_1$ and $L_2$ regularization. The experimental data is taken form \cite{eichinger2020}. In both plots, the black curve represents the experimental mean from three specimens, while the shaded areas show the mean $\pm$ the standard error of the mean. The stress is plotted in terms of $\bm{S}$ in both loading directions. Left: Stress in the principal loading direction, where the homeostatic force is decreased by $-10\%$ at $t=27$ [h]. Right: Stress in the perpendicular loading direction, which is not actively perturbed by the experimental device, though stress may still be affected indirectly.}
\label{fig:testing_cross_semibiaxial_min}
\end{figure}

\section{Discussion and current limitations}
\label{sec:limits}
The results from the previous section demonstrate that the iCANN, enhanced by homeostatic surfaces, successfully uncovers tensional homeostasis. However, as with any early stage approach, we encountered certain limitations during the discovery process and structural simulations, which we will discuss in the following section.\newline

\textbf{Discussion of activation functions.} First, we would like to discuss the choice of activation functions. As noted in Section~\ref{sec:FFN_potential}, we included the negative maximum function, though it may be unnecessary for tensional homeostasis. Alternatively, we could replace it with the absolute value function, which satisfies the pseudo potential constraints as well. The discovered weights are listed in Tables~\ref{tab:weights_stripe_abs} (stripe) and \ref{tab:weights_cross_abs} (cross).

With $L_1$ regularization, the weights are zero, except for $w_{\tau,ABS}^\phi$ in Table~\ref{tab:weights_cross_abs}. Additionally, $w_{\tau,ABS}^\phi = w_{\tau,4}^\phi$ across specimens and regularization, as the absolute value function equals the positive maximum due to the non-negative principal shear stresses.

The key result is $w_{\sigma,ABS}^\phi$ in Table~\ref{tab:weights_stripe_abs}. For our one-dimensional stress-time data, principal stresses are non-negative. However,  negative principal stress may occur in more complex deformation states, and thus, compressive stresses could contribute to the homeostatic surface, i.e., $\mathrm{abs}(\sigma_i) \geq 0$ for $\sigma_i \leq 0$. This contradicts the concept of tensional homeostasis.

Thus, while incorporating more activation functions may enhance a generic iCANN, caution is required. Sparse data can lead to non-unique or inaccurate weights, misrepresenting material behavior. To overcome this particular issue, we could have subjected the specimen to external compression. This loading scenario should teach the neural network that including the absolute value function is not allowed to contribute to the pseudo potential. However, this is not apparent from the available data.\newline

\textbf{Limitations of structural simulation.} Next, we address the unstable behavior observed in the structural simulation of the stripe specimen (Figure~\ref{fig:structural_stripe}). 
In Section~\ref{sec:results}, we derived formulas for the initial shear and bulk moduli. From these, we compute the Young's modulus, $E = \frac{9\,\kappa\,\mu}{3\,\kappa+\mu}=1.79504~\si{[\mu \N \per \mm\squared]}$, governing stress-strain relations for uniaxial loading, and Poisson's ratio, $\nu = \frac{3\,\kappa-2\,\mu}{6\,\kappa+2\,\mu} = -0.2189~[-]$, which controls lateral contraction under uniaxial deformation.
It is interesting to see that Poisson's ratio is even negative for the linear approximation in terms of the linearized theory (cf. \ref{app:moduli}).
After $t=8$ [h], we observed a strain concentration localized in the edge elements, suggesting instability. This behavior may have several causes, but we offer one primary explanation.

Achieving the homeostatic stress (Figure~\ref{fig:training_testing_stripe}), which exceeds $\sigma_{hom}\geq10~\si{[\mu \N \per \mm\squared]}$, requires a very high longitudinal strain -- under the given assumptions $\varepsilon = \frac{\sigma_{hom}}{E}$. While this explanation assumes a linear stress-strain relationship — which does not strictly apply — it highlights the core issue: The homeostatic stress is a unique material parameter, whereas the remaining parameters are non-unique in the one-dimensional case. This results in overly soft material behavior when considering structural response.

\begin{figure}[!htp]
\vspace*{-1cm}
  \centering 

  \begin{subfigure}{.48\textwidth} 
    \centering 
    \begin{tikzpicture}
      \node[inner sep=0pt] (pic) at (0,0) {\includegraphics[width=\textwidth]{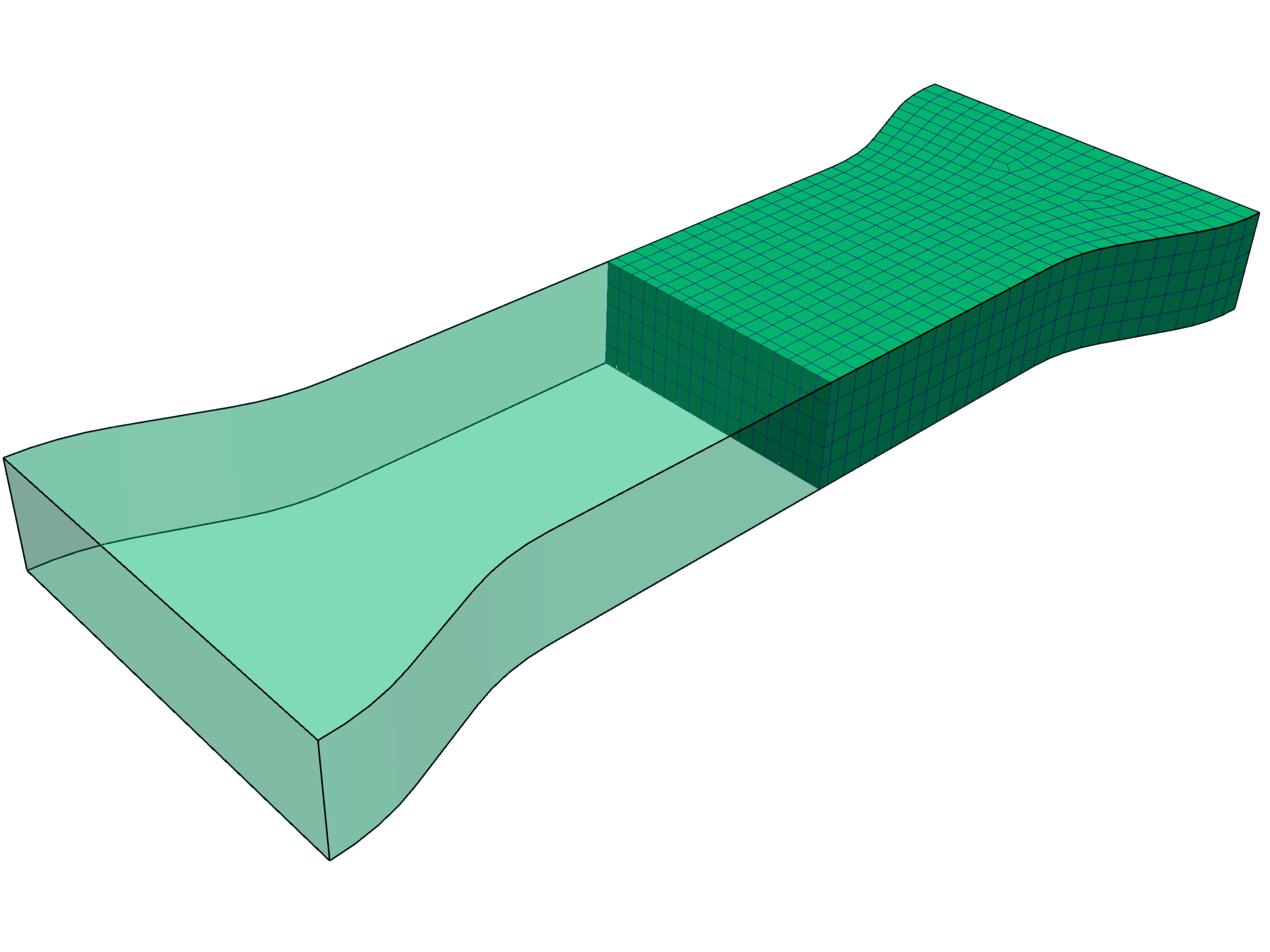}};
    \end{tikzpicture} 
    \caption{$t=0$ [h]}
  \end{subfigure}
  \begin{subfigure}{.48\textwidth} 
    \centering 
    \begin{tikzpicture}
      \node[inner sep=0pt] (pic) at (0,0) {\includegraphics[width=\textwidth]{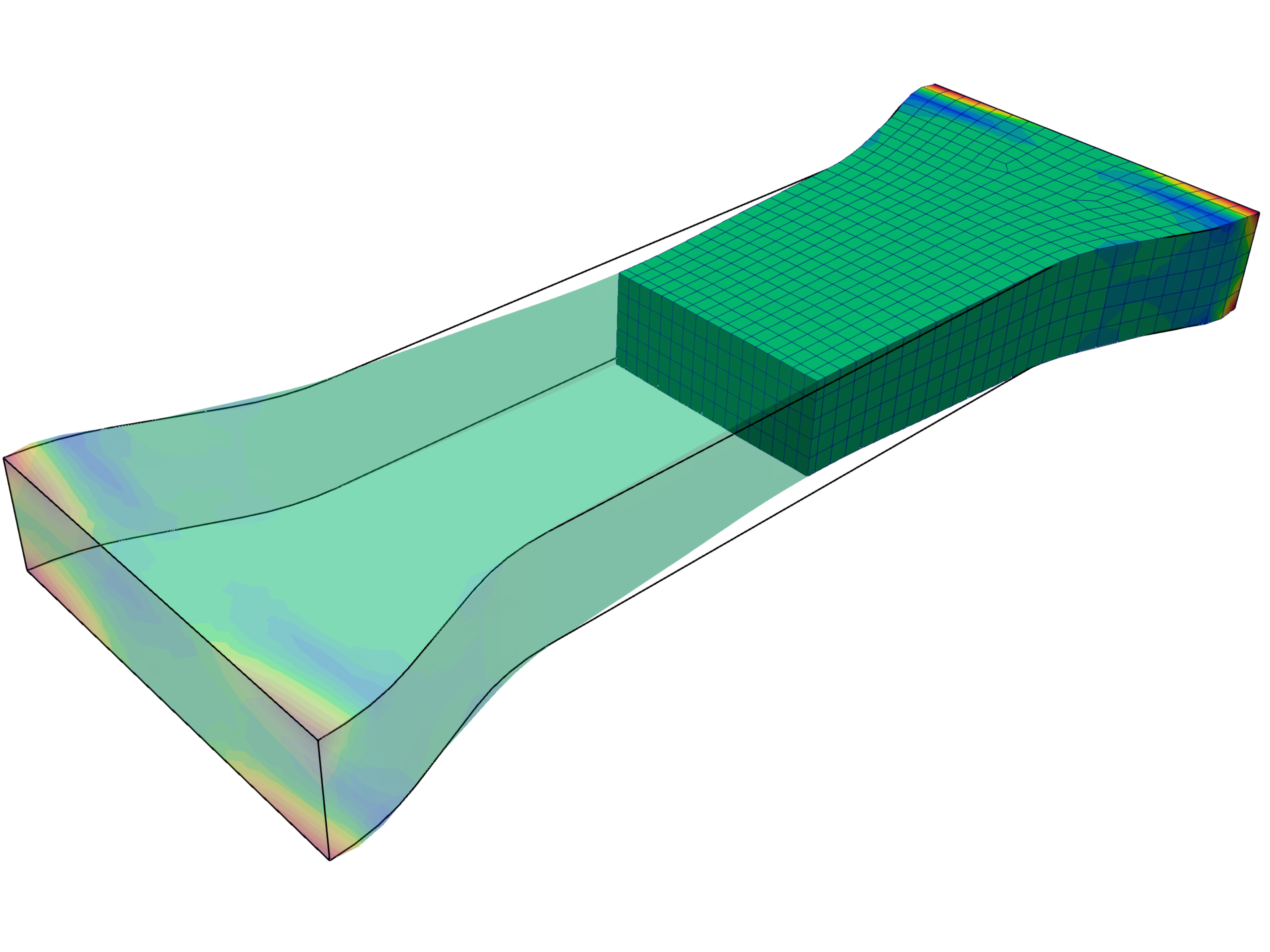}};
    \end{tikzpicture} 
    \caption{$t=6.5$ [h]}
  \end{subfigure}
  
  \begin{subfigure}{.48\textwidth} 
    \centering 
    \begin{tikzpicture}
      \node[inner sep=0pt] (pic) at (0,0) {\includegraphics[width=\textwidth]{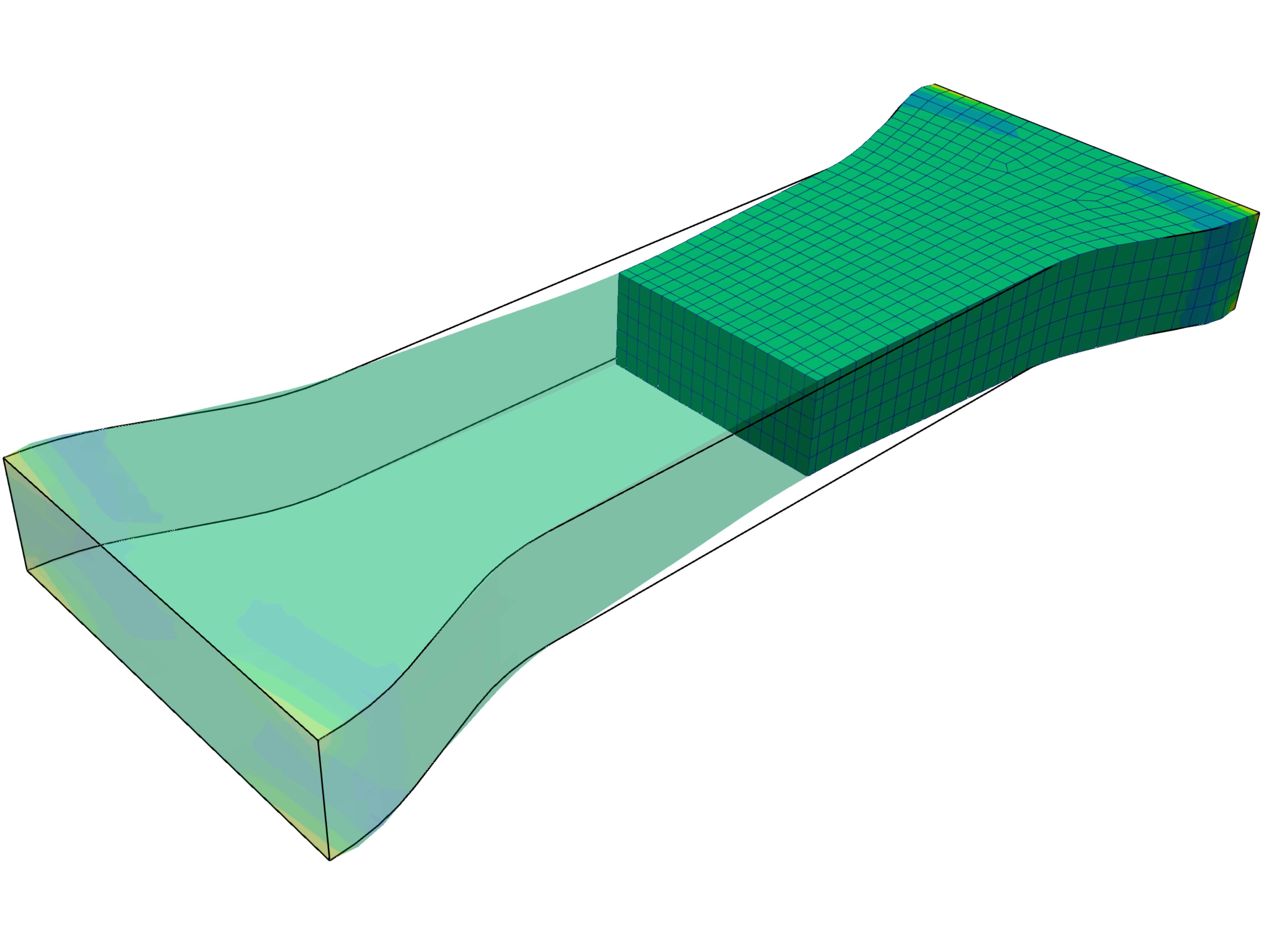}};
      \end{tikzpicture} 
    \caption{$t=8$ [h]}
  \end{subfigure}
  \begin{subfigure}{.48\textwidth} 
    \centering 
    \begin{tikzpicture}
      \node[inner sep=0pt] (pic) at (0,0) {\includegraphics[width=\textwidth]{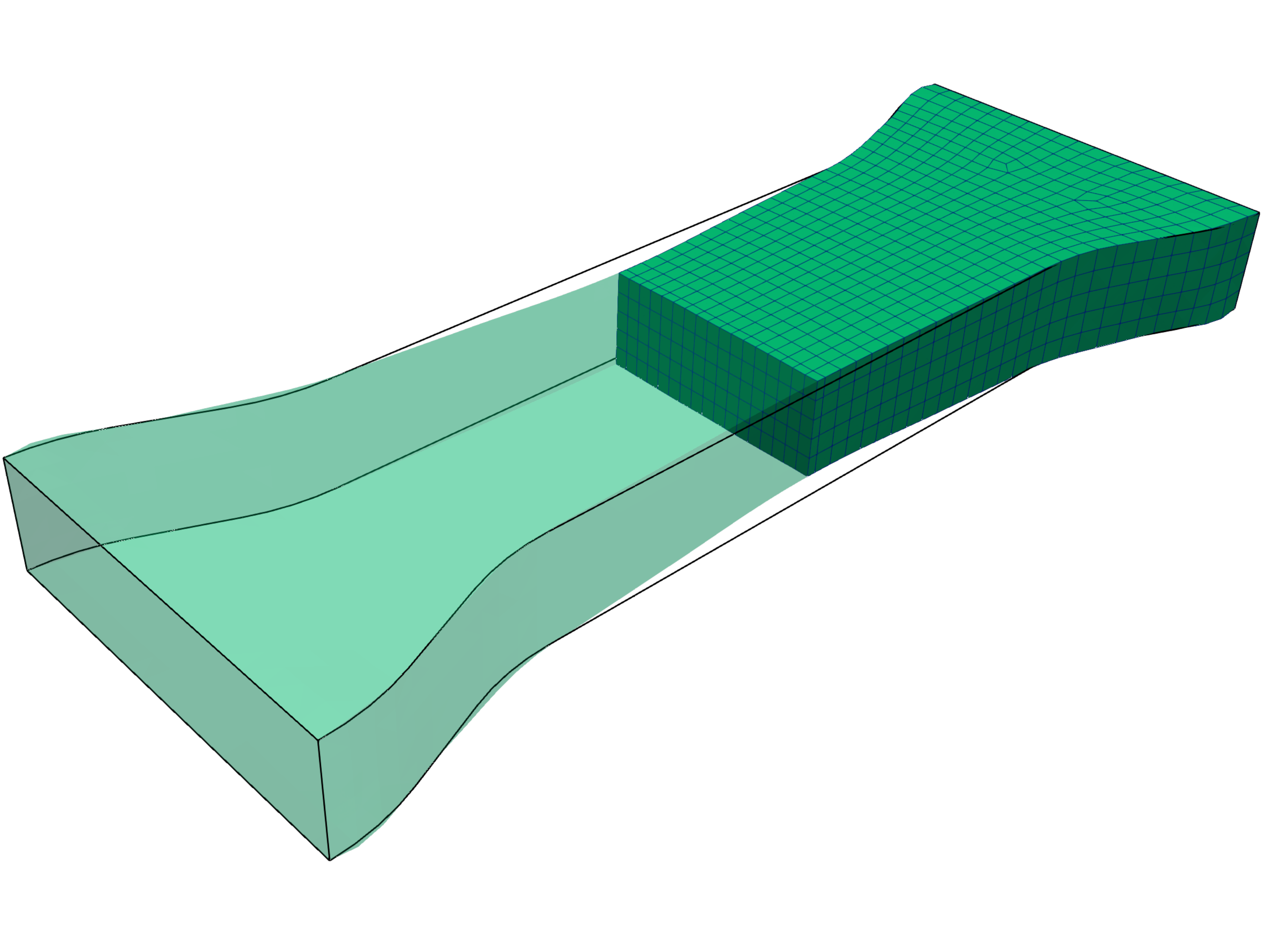}};
    \end{tikzpicture} 
    \caption{$t=17$ [h]}
  \end{subfigure}
  
  \begin{subfigure}{.48\textwidth} 
    \centering 
    \begin{tikzpicture}
      \node[inner sep=0pt] (pic) at (0,0) {\includegraphics[width=\textwidth]{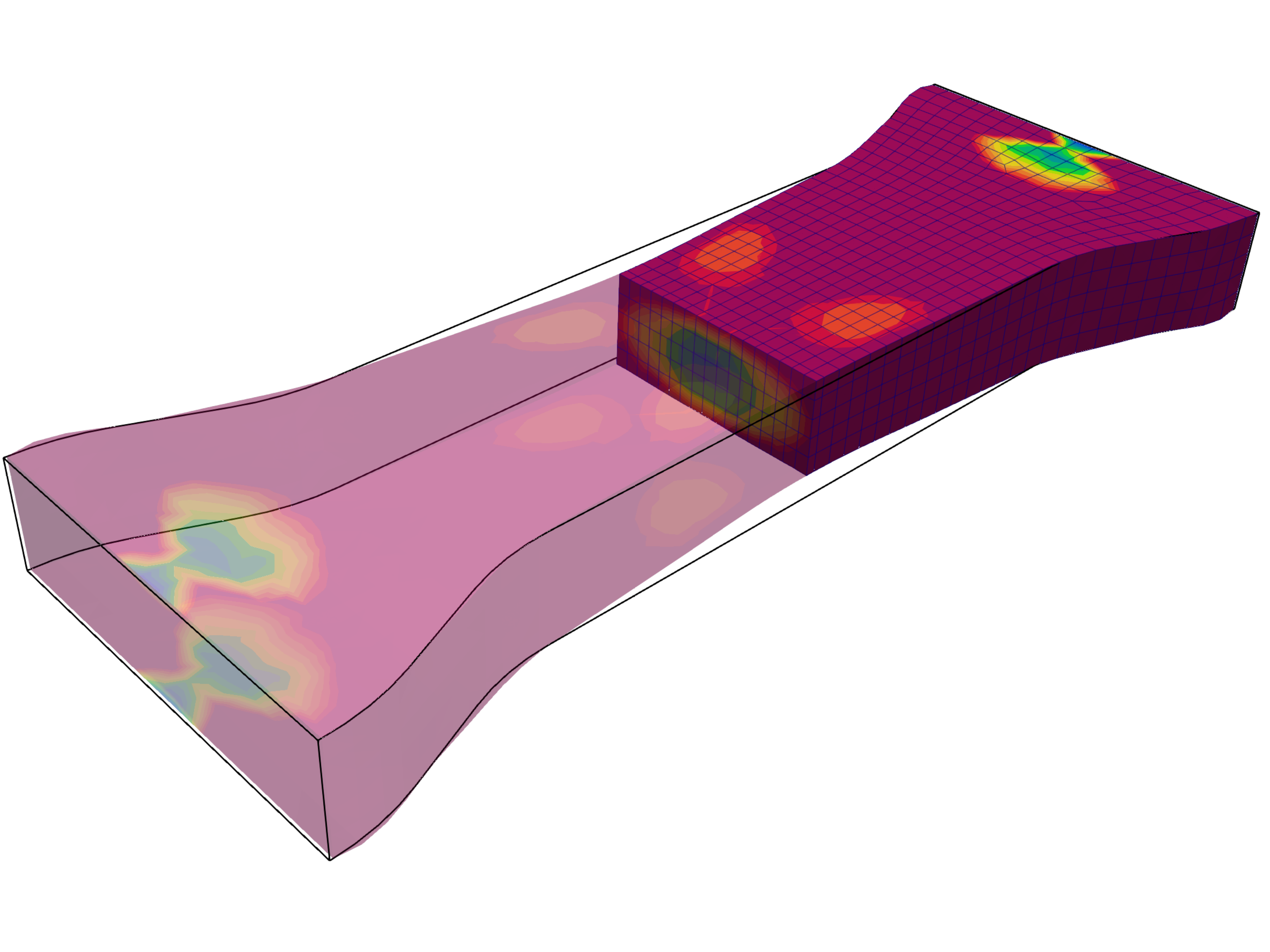}};
      \end{tikzpicture} 
    \caption{$t=17.1$ [h]}
  \end{subfigure}
  \begin{subfigure}{.48\textwidth} 
    \centering 
    \begin{tikzpicture}
      \node[inner sep=0pt] (pic) at (0,0) {\includegraphics[width=\textwidth]{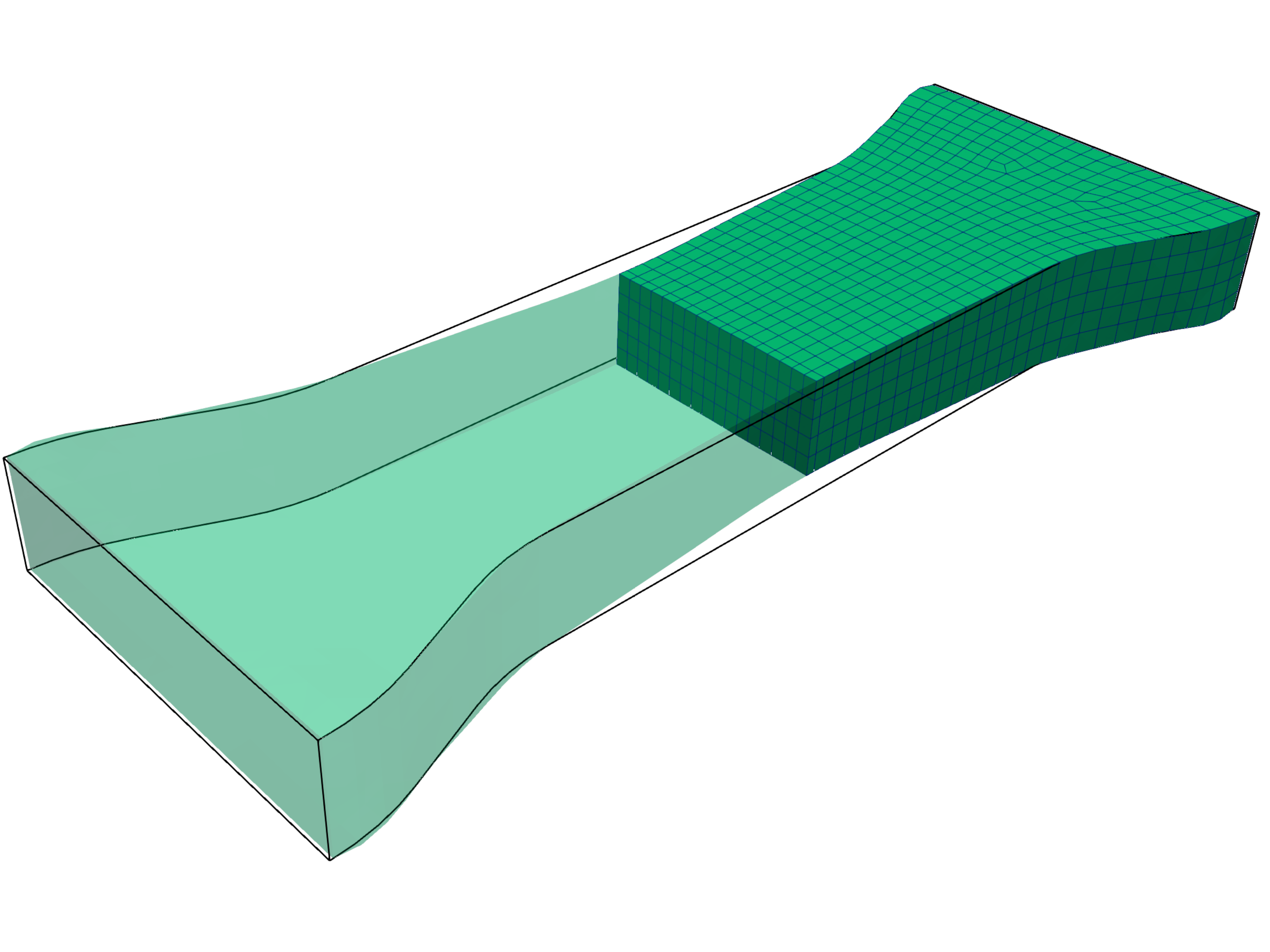}};
    \end{tikzpicture} 
    \caption{$t=27$ [h]}
  \end{subfigure}  
  
  \vspace{2mm}
  
  \begin{subfigure}{\textwidth} 
    \centering 
    \begin{tikzpicture}
      \node[inner sep=0pt] (pic) at (0,0) {\includegraphics[height=5mm, width=40mm]{figures/Damage_Step_Horizontal.pdf}};
      \node[inner sep=0pt] (0)   at ($(pic.south)+(-2.85, 0.26)$)  {$-0.0016$};
      \node[inner sep=0pt] (1)   at ($(pic.south)+( 2.85, 0.26)$)  {$\geq 0.0046$};
      \node[inner sep=0pt] (2)   at ($(pic.south)+(-1.0, -0.26)$)  {$0$};
      \node[inner sep=0pt] (d)   at ($(pic.south)+(-5.20, 0.26)$)  {$\hat{\gamma}~~\si{[\mu \N \per \mm\squared\hour]}$};
      \node[inner sep=0pt] (d)   at ($(pic.south)+( 5.20, 0.26)$)  {\hphantom{$\hat{\gamma}~~\si{[\mu \N \per \mm\squared\hour]}$}};
    \end{tikzpicture} 
  \end{subfigure}
  
  \caption{Growth multiplier, $\hat{\gamma}$, contour plots at different time steps for the adjusted model of the stripe with $w_{1,2}^\psi=39.78735$ (the remaining weights are the same as listed in Table~\ref{tab:weights_stripe}). On both edges, the specimen is fully clamped. The simulation shown corresponds to compressing the specimen. A displacement jump from zero to $u=\pm 0.101898518$ [mm] at the clamped edges is applied between $t=17$ [h] and $t=17.1$ [h]. The black lines present the undeformed shape of the specimen.} 
  \label{fig:structural_stripe_stiff}     
\end{figure}

To support our explanation, we reran the structural simulation, increasing the weight $w_{1,2}^\psi$ by a factor of $10^2$. The deformed structure, shown in Figure~\ref{fig:structural_stripe_stiff}, now exhibits a more realistic deformation and achieves a fully converged solution. Additionally, the growth multiplier is plotted, showing that once the force is perturbed in a displacement-driven manner at $t=17\ [h]$, growth and remodeling occur until homeostasis is restored, where the multiplier reaches zero.

While this does not definitively prove that our explanation is the sole cause of the instability, it demonstrates that the iCANN framework, enhanced by homeostatic surfaces, provides good qualitative and quantitative results at the material point level and reasonable results at the structural level.
As with the activation function issue, we attribute this instability to inappropriate discovered weights, likely due to the sparse data.\newline

\textbf{Limiting factors due to the assumptions made.} The eigenvalues of the elastic Mandel-like stress, $\bar{\bm{\Sigma}}$, used in the homeostatic surface equation, match those of the Kirchhoff stress tensor, $\bm{F}\bm{S}\bm{F}^T$ (see \cite{dettmer2004}). However, the eigenvalues of the relative stress (\ref{app:lagrange}) do not generally coincide with those of the Kirchhoff stress.

This leads to a key consequence: While we derived $\bm{S}$, and therefore the Kirchhoff stress, under the assumption of `directional' incompressibility — where strain and stress are zero in off-principal loading directions — this assumption does not apply to $\bar{\bm{\Sigma}}$. As a result, we cannot fully guarantee that $\bar{\bm{\Sigma}}$ represents a uniaxial stress state, which may affect the model discovery and structural simulation outcomes.
However, it is important to note that $\bar{\bm{\Sigma}}$ would represent a uniaxial stress state in reality without the Lagrange multiplier.

It is worth noting that the assumption in \ref{app:lagrange} is not intrinsic to the framework but was necessary due to sparse data for off-principal deformations. Nonetheless, we must keep this in mind when interpreting the results.\newline

\textbf{Sparsity of data sets.} In experimental mechanics, data sparsity often presents significant challenges, particularly when training standard neural networks. These networks typically require large data sets to generalize well and extract meaningful patterns. However, embedding physical principles within neural networks offers a way to mitigate this limitation. By incorporating domain-specific physics into the network’s architecture or training process, the models become better suited for interpreting sparse data sets, as they are constrained to learn physically plausible relations. This generally reduces the amount of data required for accurate predictions.

Despite this advantage, there remains a critical lower bound on sparsity. If the data becomes too sparse, the model may fail to capture essential interactions, leading to erroneous conclusions. For example, without sufficient data to inform the network about lateral contraction in materials, the model might infer incorrect relationships between weights, misrepresenting material behavior.

To address this, biaxial and semi-biaxial loadings were employed, including non-proportional loadings. However, the applied loadings appeared insufficient to uniquely identify both shear and bulk modulus. The semi-biaxial loading in particular, while intended to introduce nonlinearity, showed results almost indistinguishable from the biaxial case, indicating its unsuitability for capturing the full range of mechanical responses.

This observation points to the need for richer data sets that encompass more complex, inhomogeneous deformation states. Such data, particularly if it includes information from all three directions of loading, would provide a more robust foundation for training physics-embedded neural networks and ensuring the accurate discovery of material properties.\newline

\textbf{Model discovery at the structural level.} To address the limitations imposed by often sparse data sets in experimental mechanics, we conclude shifting towards training neural networks on structural boundary value problems. This approach has been successfully implemented in an \textit{unsupervised} fashion, as demonstrated in \cite{flaschel2021}. Specifically, we aim to represent highly complex inhomogeneous deformation states, for example, by introducing singularities such as holes in the specimen or subjecting the material to non-proportional triaxial loading scenarios. Through digital image correlation (DIC), we can capture the displacement field, while applying specific loading allows us to measure the corresponding forces.

However, this methodology entails solving boundary value problems, which often requires the use of the Finite Element Method (FEM) at each training step. Incorporating FEM into the training process would substantially increase computational costs, potentially making the approach impractical for large-scale simulations or iterative training cycles.

As an alternative, we can consider a strategy similar to the one proposed in \cite{leygue2018}. In this framework, the stress-strain fields for arbitrary specimens are precomputed in a preprocessing step. By deliberately choosing inhomogeneous deformation states, a greater variety of independent data sets can be obtained. These precomputed data sets would then be used to train neural networks directly on the stress-strain relationships, thus circumventing the need to solve the boundary value problem during training. This would significantly reduce the computational burden.

Nevertheless, one key question remains: How applicable this approach is when dealing with finite inelasticity? The nonlinear and path-dependent nature of finite inelastic deformation could challenge the precomputation of stress-strain fields. This remains an open area of investigation and could determine the viability of this alternative approach for complex, real-world materials.
\begin{table}[h]
\centering
\begin{tabular}{l | r r r}
 Weights 		& Regularized & $L_1=0.01$ & $L_2=0.001$ \\
\hline
 $w_{0,1}^\psi$ 			& No  & 1.7982913 	&   1.5855898\\
 $w_{0,2}^\psi$ 			& Yes & 	0.08166084	&   0.02207945\\
 $w_{1,1}^\psi$ 			& No  & 	-0.2862283	&   2.0337853\\
 $w_{1,2}^\psi$ 			& Yes & 	0			&   0.08938348\\
 $w_{\sigma,ABS}^\phi$		& Yes &	0			&   0.00010657\\
 $w_{\sigma,2}^\phi$		& Yes &	0			&   0\\ 
 $w_{\sigma,3}^\phi$		& No  &	7.898215e-07	&   3.9903475e-07\\
 $w_{\sigma,4}^\phi$		& Yes &	0.02767334	&   0.00130172\\
 $w_{\tau,ABS}^\phi$		& Yes &	0			&   0.02694412\\
 $w_{\tau,2}^\phi$		& Yes &	0			&   0\\ 
 $w_{\tau,3}^\phi$		& No  &	1.1955261e-08	&   1.10272524e-07\\
 $w_{\tau,4}^\phi$		& Yes &	0			&   0.02694412\\ 
 $\hat{w}_\eta$				& Yes &	0.44342846\tablefootnote{regularized with $L_2=0.001$}		& 0.31261802
\end{tabular}
\caption{Discovered weights for the stripe specimen (uniaxial stress state) with $L_1$ and $L_2$ regularization. We replaced the $\mathrm{max}\left(-(\bullet),0\right)$ activation function by the $\mathrm{abs}\left(\bullet\right)$ function with the corresponding weights $w_{\sigma,ABS}^\phi$ and $w_{\tau,ABS}^\phi$. The second column, `Regularized', indicates whether the regularization factor associated with the weight is zero (`No') or if regularization, $L_1$ or $L_2$, is applied (`Yes').}
\label{tab:weights_stripe_abs}
\end{table}
\begin{table}[!htp]
\centering
\begin{tabular}{l | r r r}
 Weights 		& Regularized & $L_1=0.01$ & $L_2=0.001$ \\
\hline
 $w_{0,1}^\psi$ 			& No &   	2.1491084		&   2.26735\\
 $w_{0,2}^\psi$ 			& Yes & 	0.39225546		&   0.3615928\\
 $w_{1,1}^\psi$ 			& No & 	-1.6436962		&   -0.35730883\\
 $w_{1,2}^\psi$ 			& Yes & 	0				&   0.00483087\\
 $w_{\sigma,ABS}^\phi$		& Yes &	0				&   0\\
 $w_{\sigma,2}^\phi$		& Yes &	0				&   0\\ 
 $w_{\sigma,3}^\phi$		& No &	4.9472845e-07	&   8.272873e-07\\
 $w_{\sigma,4}^\phi$		& Yes &	0.01457978		&   0.0146279\\
 $w_{\tau,ABS}^\phi$		& Yes &	0.00533062		&   0.00149194\\
 $w_{\tau,2}^\phi$		& Yes &	0				&   0\\ 
 $w_{\tau,3}^\phi$		& No &	7.6053965e-07	&   1.7517864e-07\\
 $w_{\tau,4}^\phi$		& Yes &	0.00533062		&   0.00149194\\ 
 $\hat{w}_\eta$				& Yes &	0.2940081\tablefootnote{regularized with $L_2=0.001$}		& 0.28590617
\end{tabular}
\caption{Discovered weights for the cross specimen (biaxial stress state) with $L_1$ and $L_2$ regularization. We replaced the $\mathrm{max}\left(-(\bullet),0\right)$ activation function by the $\mathrm{abs}\left(\bullet\right)$ function with the corresponding weights $w_{\sigma,ABS}^\phi$ and $w_{\tau,ABS}^\phi$. The second column, `Regularized', indicates whether the regularization factor associated with the weight is zero (`No') or if regularization, $L_1$ or $L_2$, is applied (`Yes').}
\label{tab:weights_cross_abs}
\end{table}
%
%
\section{Conclusion and outlook}
\label{sec:outlook}
In conclusion, this work represented a step forward in uncovering the mechanisms of tensional homeostasis at the material point level by enhancing the iCANN framework for growth and remodeling. Through the integration of homeostatic surfaces into neural network architectures, we not only taught networks to grasp this complex theory but also demonstrated the capability to predict the state of homeostasis under finite strain. Importantly, our framework is rooted in thermodynamic consistency, ensuring that the neural networks generate physically sound predictions beyond the training regime, which marks a notable achievement in constitutive modeling of biological tissues.

Moreover, we successfully implemented equality constraints essential for maintaining the integrity of homeostatic surfaces within neural networks, establishing a solid theoretical foundation for modeling tensional homeostasis. This novel architecture has the potential to redefine how we approach highly nonlinear material behaviors.

Despite these advancements, we encountered instabilities when extending the framework to the structural level. Challenges such as the choice of activation functions and the discovery of weights at the material point level emerged, particularly due to the sparse datasets used to capture highly nonlinear material behaviors. While our approach was able to qualitatively match expected structural behavior to some extent, the sparsity of the data limited its full potential.

Looking ahead, future work should not only emphasize computational improvements but also prioritize appropriate specimen design. We believe that successful discovery will require a combined effort from both computational and experimental perspectives. The simplicity of the current architecture, designed for comprehensiveness, presents opportunities for further development to integrate higher-order nonlinearities within the homeostatic surface model.
Additionally, it is crucial to extend the network to account for more complex interactions between input variables, such as hydrostatic pressure. Equally important is the incorporation of uncertainty into the model discovery process. By systematically addressing uncertainties in the data and model predictions, we can enhance the robustness and reliability of the framework.
By combining these computational advances with more sophisticated experimental designs, we can improve the precision of our insights and the accuracy of models depicting complex material behavior. 
This can pave the way for more precise insights and accurate models of complex material behavior, giving us a deeper understanding of the underlying mechanisms.

\appendix
\section{Appendix} 
\subsection{Lagrange multiplier}
\label{app:lagrange}
To enforce zero strain and stress in the off-principal axes direction, we add a Lagrange term to the energy, $\psi_{\lambda}=-p\left(\mathrm{tr}\left(\bm{C}\bm{M}\right)-1\right)$, where $\bm{M}$ denotes a structural tensor.
This tensor is aligned with the Cartesian axes, i.e. $\bm{M}=\bm{e}_2\otimes\bm{e}_2$ or $\bm{M}=\bm{e}_3\otimes\bm{e}_3$.
First, we exploit that
\begin{equation}
	\psi_{\lambda}=-p\left(\mathrm{tr}\left(\bm{C}\bm{M}\right)-1\right) = -p\left(\mathrm{tr}\left(\bm{U}_g\bar{\bm{C}}_e\bm{U}_g\bm{M}\right)-1\right).
\end{equation}
Since both uniaxial and biaxial loadings are coaxial loading scenarios, the principal axes of $\bm{C}$ do not rotate, and thus, both $\bar{\bm{C}}_e$ and $\bm{U}_g$ are aligned with these principal axes and are coaxial (cf \cite{itskov2004}).
Thus, we obtain 
\begin{equation}
	\psi_{\lambda}\left(\bar{\bm{C}}_e,\bm{C}_g,\bm{M}\right) = -p\left(\mathrm{tr}\left(\bar{\bm{C}}_e\bm{C}_g\bm{M}\right)-1\right).
\end{equation}
Noteworthy, this is only true for the cases we investigate in this contribution.
With this at hand, the rate of the energy reads
\begin{equation}
	\dot{\psi}_\lambda = \frac{\partial\psi_\lambda}{\partial\bar{\bm{C}}_e} : \dot{\bar{\bm{C}}}_e + \frac{\partial\psi_\lambda}{\partial\bm{C}_g} : \dot{\bm{C}}_g
\end{equation}
since the rate of $\bm{M}$ is equal to zero.
We rearrange the latter equation, which yields
\begin{equation}
	\underbrace{2\,\bm{U}_g^{-1}\frac{\partial\psi_\lambda}{\partial\bar{\bm{C}}_e}\bm{U}_g^{-1}}_{=:\bm{S}_\lambda} : \frac{1}{2}\dot{\bm{C}} - \left( \underbrace{2\,\bar{\bm{C}}_e\frac{\partial\psi_\lambda}{\partial\bar{\bm{C}}_e}}_{=:\bar{\bm{\Sigma}}_\lambda} - \underbrace{2\,\frac{\partial\psi_\lambda}{\partial\bm{C}_g}\bm{C}_g}_{=:\bar{\bm{X}}_\lambda}\right) : \bar{\bm{L}}_g
\end{equation}
Next, we investigate the relative stress, $\bar{\bm{\Sigma}}_\lambda - \bar{\bm{X}}_\lambda$, for the particular choice of $\psi_\lambda$
\begin{align}
	\bar{\bm{\Sigma}}_\lambda &= 2\,\bar{\bm{C}}_e\left(\frac{-p}{2}\left(\bm{C}_g\bm{M} + \bm{M}\bm{C}_g\right)\right) \\
	\bar{\bm{X}}_\lambda &= 2\,\left(\frac{-p}{2}\left(\bar{\bm{C}}_e\bm{M} + \bm{M}\bar{\bm{C}}_e\right)\right)\bm{C}_g.
\end{align}
Consequently, the relative stress reads
\begin{equation}
	\bar{\bm{\Sigma}}_\lambda - \bar{\bm{X}}_\lambda = -p\left(\bar{\bm{C}}_e\bm{C}_g\bm{M} - \bm{M}\bar{\bm{C}}_e\bm{C}_g\right)
\end{equation}
which is generally not symmetric.
However, as exploited above, all tensors are coaxial in our cases.
Hence, the relative stress resulting from the Lagrange energy is zero.
Therefore, we do not have to consider its contribution in the growth potential $\hat{\phi}$.
In total, we only need to consider the contribution to the second Piola-Kirchhoff Stress 
\begin{equation}
	\bm{S}_\lambda = -p\,\bm{U}_g^{-1}\left( \bm{C}_g\bm{M} + \bm{M}\bm{C}_g \right)\bm{U}_g^{-1} = -2\,p\, \bm{M}
\end{equation}
where the last equation was again obtained under the consideration of coaxility.
Finally, the Lagrange multiplier, $p$, is calculated such that the stress in the direction of $\bm{M}$ is zero.
%
\subsection{Consistency condition with linearized theory}
\label{app:moduli}
In order to interpret our discovered weights in terms of the (initial) shear, $\mu$, and bulk, $\kappa$, moduli, we can employ that our models must be consistent with the linearized theory of elasticity (see \cite{ogdenbook}).
This states the following
\begin{equation}
	\left. \frac{\partial^2\psi}{\partial\xi_i\partial\xi_j} \right|_{\xi_1=\xi_2=\xi_3=1} = \kappa - \frac{2}{3}\,\mu + 2\,\mu\,\delta_{ij} \quad i,j \in \left[1,2,3\right]
\end{equation}
where $\delta_{ij}$ denotes the Kronecker delta and $\xi_i^2 = \lambda_i$ with $\lambda_i$ being the eigenvalues of the (elastic) right Cauchy Green tensor (cf. Section~\ref{sec:Helmholtz}).
With this equation at hand, we end up with we following two linear independent equations for our designed energy in Equation~\eqref{eq:ffn_energy}
\begin{align}
	\kappa + \frac{4}{3}\,\mu &= 4\, w_{0,2}^\psi\, \left(w_{0,1}^\psi\right)^2 + \frac{8}{3}\, w_{1,2}^\psi\, \left(w_{1,1}^\psi\right)^2 \\
	\kappa - \frac{2}{3}\,\mu &= 4\, w_{0,2}^\psi\, \left(w_{0,1}^\psi\right)^2 - \frac{4}{3}\, w_{1,2}^\psi\, \left(w_{1,1}^\psi\right)^2.
\end{align}
From this set of equations, we can deduce relations between our weights and the material parameters known from classical theory, i.e.,
\begin{equation}
	\kappa = 4\, w_{0,2}^\psi\, \left(w_{0,1}^\psi\right)^2, \quad \mu = 2\, w_{1,2}^\psi\, \left(w_{1,1}^\psi\right)^2.
\end{equation} 
\section{Declarations}
%
\subsection{Acknowledgements}
First of all, the authors would like to express their gratitude to Daniel Paukner and Christian Cyron for providing the experimental data.
The authors are also grateful for fruitful discussions with Jannick Kehls on the numerical implementation.
Further, Hagen Holthusen and Tim Brepols gratefully acknowledge financial support of the projects 417002380 and 453596084 by the Deutsche Forschungsgemeinschaft.
In addition, Kevin Linka is supported by the Emmy Noether Grant 533187597 and project 465213526 by the Deutsche Forschungsgemeinschaft.
This work was supported by the NSF CMMI Award 2320933 Automated Model Discovery for Soft Matter and by the ERC Advanced Grant 101141626 DISCOVER to Ellen Kuhl.
%
%
\subsection{Conflict of interest}
The authors of this work certify that they have no affiliations with or involvement in any organization or entity with any financial interest (such as honoraria; participation in speakers’ bureaus; membership, employment, consultancies, stock ownership, or other equity interest; and expert testimony or patent-licensing arrangements), or non-financial interest (such as personal or professional relationships, affiliations, knowledge or beliefs) in the subject matter or materials discussed in this manuscript.
%
\subsection{Availability of data and material}
Our data used for training is accessible to the public at \url{https://doi.org/10.5281/zenodo.13946282}
%
\subsection{Code availability}
Our source code and examples of the iCANN implementation in \textit{Keras/TensorFlow} as well as a material subroutine implemented in \textit{FORTRAN} are accessible to the public at \url{https://doi.org/10.5281/zenodo.13946282}.
We followed the approach of \cite{korelc2014} for the computation of the exponential map and \cite{hudobivnik2016} to compute the matrix square root.
%
\subsection{Contributions by the authors}
\textbf{Hagen Holthusen:} Conceptualization, Methodology, Software, Validation, Formal analysis, Investigation, Data Curation, Writing - Original Draft, Writing - Review \& Editing, Visualization, Funding acquisition\\
\textbf{Tim Brepols:} Methodology, Writing - Original Draft, Writing - Review \& Editing, Supervision, Funding acquisition\\
\textbf{Kevin Linka:} Methodology, Software, Writing - Original Draft, Writing - Review \& Editing, Funding acquisition\\
\textbf{Ellen Kuhl:} Funding acquisition, Methodology, Writing – original draft, Writing – review \& editing.
%
\subsection{Statement of AI-assisted tools usage}
This document was prepared with the assistance of OpenAI's ChatGPT, an AI language model. ChatGPT was used for language refinement. The authors reviewed, edited, and take full responsibility for the content and conclusions of this work.


\end{document}